\documentclass[lettersize,journal]{IEEEtran}

\usepackage{amsmath,amsfonts}
\usepackage[linesnumbered,ruled,vlined]{algorithm2e}%
\usepackage{amsthm,amssymb}
\usepackage{array}
\usepackage{url}
\usepackage{verbatim}
\usepackage{graphicx}
\usepackage{cite}
\usepackage{color}
\usepackage{subfigure}
\usepackage{textcomp}
\newcommand{\algcomments}[1]{\textit{\texttt{\# #1}}}
\usepackage{setspace}
\theoremstyle{remark}
\newtheorem{remark}{Remark}%[section]

\SetKwInOut{Required}{Required}
\SetKw{Break}{break}

\begin{document}

\title{Autonomous Robotic Ultrasound System for Liver Follow-up Diagnosis: Pilot Phantom Study}

\author{Tianpeng Zhang$^{\dagger}$, Sekeun Kim$^{\dagger}$, Jerome Charton$^{\dagger}$, Haitong Ma, Kyungsang Kim$^*$, Na Li, Quanzheng Li 
        % <-this % stops a space
\thanks{T. Zhang, H. Ma and Na Li are with the School of Engineering and Applied Sciences, Harvard University, Cambridge, MA 02138, United States. Email: \texttt{tzhang@g.harvard.edu, nali@seas.harvard.edu.} They are supported under NSF AI institute:
2112085.}
\thanks{S. Kim, J. Charton, K. Kim and Q. Li are with the Center for Advanced Medical Computing and Analysis (CAMCA), Massachusetts General Hospital and Harvard Medical School, Boston, MA 02114, United States. Email: \texttt{SKIM207@mgh.harvard.edu, JCHARTON@mgh.harvard.edu, kkim24@mgb.org, Li.Quanzheng@mgh.harvard.edu}.}% <-this % stops a space
\thanks{$^{\dagger}$: contributed equally}
%\thanks{Manuscript received April 19, 2021; revised August 16, 2021.}
}



\maketitle

\begin{abstract}
The paper introduces a novel autonomous robot ultrasound (US) system targeting liver follow-up scans for outpatients in local communities. Given a computed tomography (CT) image with specific target regions of interest, the proposed system carries out the autonomous follow-up scan in three steps: (i) initial robot contact to surface, (ii) coordinate mapping between CT image and robot, and (iii) target US scan. Utilizing 3D US-CT registration and deep learning-based segmentation networks, we can achieve precise imaging of 3D hepatic veins, facilitating accurate coordinate mapping between CT and the robot. This enables the automatic localization of follow-up targets within the CT image, allowing the robot to navigate precisely to the target's surface. Evaluation of the ultrasound phantom confirms the quality of the US-CT registration and shows the robot reliably locates the targets in repeated trials. The proposed framework holds the potential to significantly reduce time and costs for healthcare providers, clinicians, and follow-up patients, thereby addressing the increasing healthcare burden associated with chronic disease in local communities.
\end{abstract}


\section{Introduction}
% Level of unraveling:
% \begin{itemize}
%     \item Ultrasound imaging
%     \item Robotic Ultrasound
%     \item Autonomous Robotic Ultrasound, with AI techniques. 
%     \item Probe Positioning in Autonomous Robotic Ultrasound with pre-operative images
%     \item Conventional approach to pre-op to US registration
%     \item Segmentation-based registration.
%     \item Our contribution: auto robotic ultrasound imaging for liver based on hepatic vein segmentation.
% \end{itemize}

% \tianpeng{Make two diagrams: 1) The connection between the hardware components, robot, US probe, depth camera, PC, etc. 2) The flow chart explaining the pipeline.}

\IEEEPARstart{U}ltrasound (US) is extensively utilized in clinics due to its minimal invasiveness and portability, compared to other medical imaging modalities, such as radiography (X-ray), CT, and MR. The portability of ultrasound makes it widely employed as a point-of-care (POC) modality in emergency scenarios outside of the hospital, offering cost-effectiveness, convenience, ready availability, and absence of radiation exposure \cite{saraogi2015lung}.
Within the existing POC framework, bedside ultrasound imaging plays a crucial role in real-time monitoring and physician support. However, the data quality POC ultrasound varies greatly even when operated by the same clinician. This variability is further compounded by differences in clinician skill levels, presenting challenges in reliable ultrasound imaging \cite{berg2006operator,jiang2020automatic,sonko2022machine}.

Two solutions have emerged to address the lack of skilled clinicians for POC ultrasound outside of the hospital. The first solution involves integrating artificial intelligence (AI), including machine learning (ML) and deep learning (DL), to assist less skilled healthcare providers in conducting diagnostic US scans. For example, Kim {\em et al.} \cite{kim2023point} recently developed an autonomous AI framework for diagnosing pneumothorax, which includes quality assurance, region of interest (ROI) extraction, and diagnosis. This framework follows a step-wise workflow similar to skilled clinicians' operation, enabling less skilled healthcare providers to accurately scan regions under AI guidance.
Another solution involves employing a \textbf{\textit{Robotic Ultrasound System (RUSS)}} \cite{von2021medical,li2023overview,jiang2023robotic} to replace the manual operation of ultrasound probes with robotic manipulators. An automated RUSS allows practitioners initiate US scans with simple and straightforward instructions to the robot, such as specifying a ROI to cover \cite{graumann2016robotic} or a scanning path \cite{hennersperger2016towards} based on CT/MRI images, after which the robot performs the desired US scan with consistency and precision. The past decade has seen RUSS integrated with various AI techniques, including Gaussian mixture models \cite{mylonas2013autonomous}, deep segmentation networks \cite{ramalhinho2022deep}, and deep reinforcement learning \cite{ning2021autonomic}, to enhance US image analysis and robot control components. This field is moving towards developing autonomous RUSS that are able to produce high-quality ultrasound scans and operate fully independent of skilled sonographers.

\begin{figure*}[t]
    \centering
     \includegraphics[width = 0.9\linewidth]{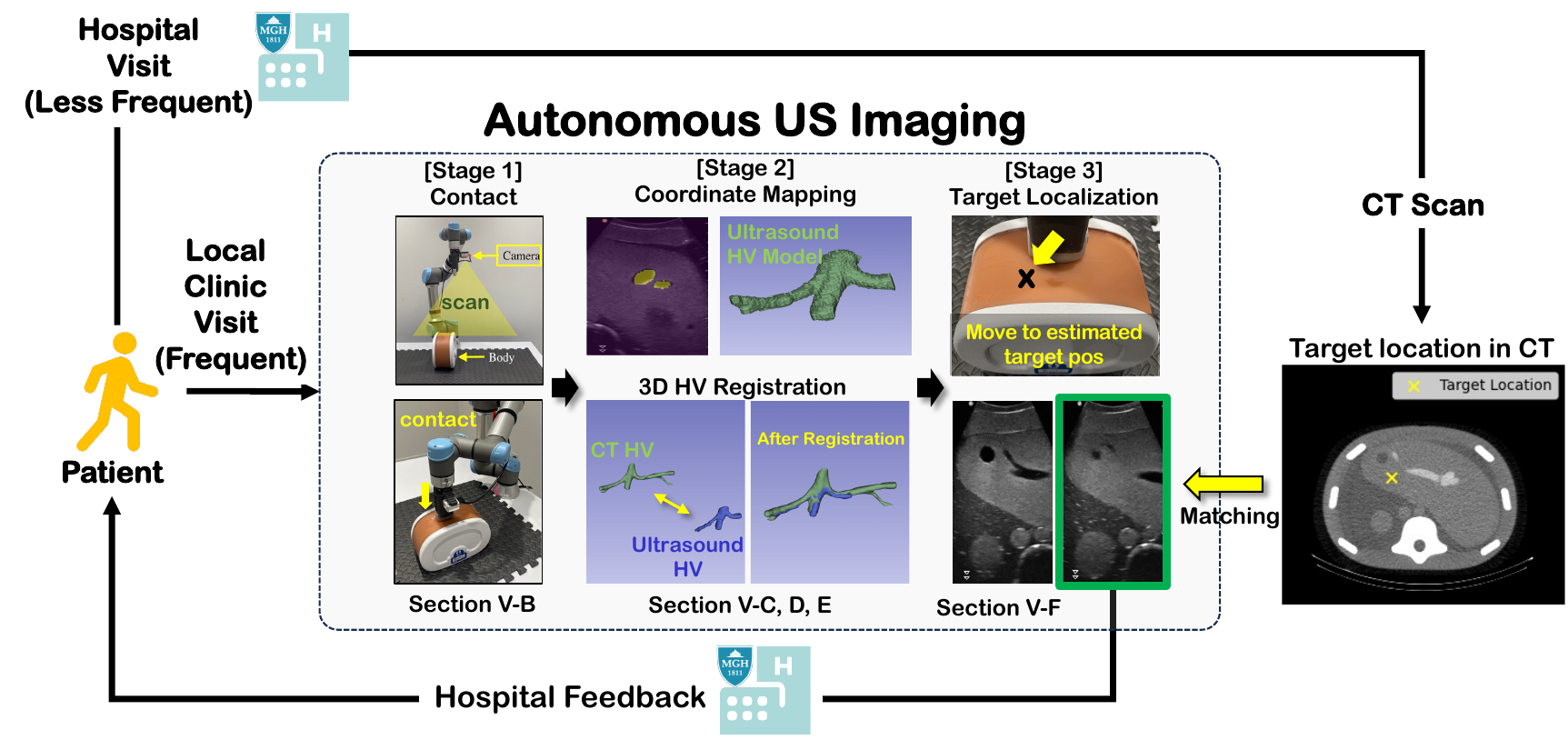}
    \caption{Our proposed robot system conducts ultrasound scans on patients for intermediate follow-up examinations at local clinics. We assume that the CT scan data is provided by the hospital. The autonomous US imaging pipeline involves initial contact, robot coordinate mapping, segmentation/registration of the hepatic vein (HV), image localization integrated with robot control, and so on. Deep segmentation networks assist the coordinate mapping and target localization. }
   \label{fig:big-system-semantics}
\end{figure*}

% \begin{figure}[htp]
%     \centering
%      \includegraphics[width = 0.75\linewidth]{figures/system_diagrams/ClinicalProcedureV3.pdf}
%     \caption{The Clinical Procedure.}
%     \label{subfig:clinical-procedure}
% \end{figure}

% The main goal of this work develop a robotic ultrasound system that can deliver substantial clinical benefits by reducing time and costs for healthcare providers, clinicians, and patients. Our attention is particularly directed towards
One important use case for autonomous RUSS is for patients requiring \textbf{\textit{follow-up}} ultrasound scans. The patients may visit the hospital for comprehensive CT/MR scans sporadically, and visit their local clinics for frequent follow-up US scans between hospital visits. In the context of a follow-up scan, we assume there are regions of interest (ROIs) specified in CT or MR images of the patient that require regular monitoring. More specifically, our clinical focus is on liver follow-up scans, prompted by the growing prevalence of nonalcoholic fatty liver disease (NAFLD). The annual cost of care per NAFLD patient with private insurance stands at \$7,804 for a new diagnosis and \$3,789 for long-term management in 2018 \cite{allen2018healthcare}. Recognizing the escalating healthcare burden associated with NAFLD, the autonomous RUSS arises as a potential solution that delivers efficient and cost-effective liver follow-up scans.

In this work, we propose an autonomous robotic ultrasound system integrated with AI models for effective and high-quality follow-up liver ultrasound scans, where we employ hepatic veins (HV) as the landmark features for liver navigation and train deep segmentation networks to identify HV structures in US images. This work is a pilot phantom study that lays down the foundation for real-patient applications in the future. Fig. \ref{fig:big-system-semantics} illustrates how our robotic US system operates as part of the typical clinical procedure: the patient sporadically visits the hospital for CT scans, while regularly attends our autonomous US imaging at the local clinic and receives hospital feedback remotely. During the local clinic visits, our system carries out autonomous US imaging in three stages. The first stage is to make contact with the body guided by the RGB-D camera view from the robot's end-effector. The second stage focuses on coordinate mapping, where the system locates the HV using the deep segmentation models, acquires a sequence of US images to form a 3D vessel model of HV in the physical space, and uses 3D registration between the 3D HV images from US and CT to compute a coordinate mapping between the CT image and physical space. The coordinate mapping provides an estimate of the target location in physical space. The third stage utilizes the estimated target coordinates and slice-matching to accurately align the US probe with the CT slice containing the target, after which the desired US images of the target location are captured. Our method pioneers the effort of leveraging deep segmentation networks in RUSS to assist real-time vessel segmentation for autonomous US probe navigation. Our method is also among the first to integrate segmentation-based 3D registration method with an autonomous US imaging pipeline, overcoming modality differences between US and CT images. Experiments on a US phantom evaluate the system. Results demonstrate that our system achieves high-quality registration and can auto-locate and capture US images for targets in CT images with a high success rate across 500 trials.

\section{Related Works}

\subsection{Medical Robotic Systems}
Robotic systems have played a crucial role in driving the impressive progress of medical technology in recent decades. Various models are developed to assist or even replace medical experts in challenging environments. Notable examples include the da Vinci Robotic Surgical System developed by Intuitive Surgical in Mountain View, California, and the Zeus Robotic Surgical System by Computer Motion in Goleta, California. These systems have been successfully integrated into surgical practices across different fields. 

In particular, robotic US system (RUSS) has shown promise in various clinical scenarios, such as diagnostic imaging\cite{nakadate2010implementation,mustafa2013development}, 3D bio-metrics\cite{merouche2015robotic,virga2016automatic,kojcev2017reproducibility}, and surgical assistance\cite{chatelain20153d,kojcev2016dual} in the past two decades. A great number of tele-operated RUSS are proposed to allow the US scanning be operated through remote control\cite{salcudean1999robot,essomba2012design}, thus reducing the stress factor on US technicians. However, tele-operated RUSS is limited as it still requires the manual control of a sonographer. 

In recent years, fully autonomous robotic US systems have been developed to offer precise and repeatable US imaging results without the manual operation of a sonographer. The early research \cite{mustafa2013development} segments the abdominal region and calculates the epigastric region using image features, allowing the robot to locate the region of interest and execute a pre-programmed simple scanning motion. Recent advancements enable technicians to only specify scanning paths \cite{hennersperger2016towards} or volumes of interest\cite{graumann2016robotic} in tomographic images (CT or MRI), while the robot completes the specified scans autonomously. Some systems can even auto-locate and scan specific structures without anatomical information, although they are often restricted to specific organs \cite{virga2016automatic,von2020robotized}. 
It worth mentioning that various AI techniques such as Gaussian mixture models \cite{mylonas2013autonomous} and deep reinforcement learning \cite{ning2021autonomic} have been employed to develop autonomous US scanning policies. Particularly, deep segmentation has emerged as a pivotal component in recent studies, serving as the foundation for controlling the positioning of US probes \cite{von2020robotized,jiang2021autonomous}. 

In this work, we propose an autonomous robotic US system, with the focus being performing autonomous US scanning using a CT image guidance. Our control pipeline utilizes deep segmentation networks for locating structures of interest within the body, implementing probe positioning strategies, and facilitating US-CT registration. Experimental results demonstrate that our system is able to reliably find and image target locations specified in the CT, without requiring any manual control from human operators.

\subsection{Image Registration}
Medical tomographic images, such as CT, PET and MRI, are widely used as the ground-truth reference about the patient's anatomy or physiology. Many existing systems are designed to follow a specific path or cover a region of interest within 3D tomographic images \cite{hennersperger2016towards,graumann2016robotic}. The key to their success lies in computing a high-quality coordinate mapping between the CT image and the physical world, which allows the robot to accurately position the US probe according to the specified path or regions. 
Two common approaches are used to compute such a mapping. The first approach involves using an RGB-D camera to capture the patient's surface in physical space and then registering it with the patient's surface in the image \cite{hennersperger2016towards,graumann2016robotic}. The second approach is to directly register the US images with the tomographic images by matching internal features within the patient's body \cite{zettinig2016toward,langsch2019robotic}. The second method holds the potential for achieving superior registration as it can access richer and more diverse features. However, aligning ultrasound (US) images with a different image modality in tomographic images, such as CT or MRI, poses a significant challenge. The $LC^2$ method \cite{fuerst2014automatic} aims to align images of different modalities by optimizing similarity metrics between them to find the spatial registration that minimizes differences. However, $LC^2$ is computationally expensive due to the need for global optimization to find the transformation $T$, and could perform unsatisfactorily when dealing with significantly different modalities. 

Segmentation-based registration methods, such as those proposed by \cite{haque2016automated, ramalhinho2022deep, he2023robust}, offer a promising approach to address the disparity in image modalities. By converting ultrasound (US) and pre-operative images into segmentation maps, these methods achieve high-quality registration by matching segmented features rather than original pixel values. Although this approach requires additional effort to annotate segments in both images, the resulting registration is more robust and reliable than similarity metric based methods like $LC^2$. 

In our work, we adopt segmentation-based registration to compute the coordinate mapping between the 3D CT image and the physical robot. We fully automate the procedure of 3D US image acquisition for HV, and compute the US-CT registration based on the segmented 3D HV structure in these two modalities. To our knowledge, our system is the first to integrate segmentation-based registration into an autonomous robotic US pipeline.

\section{Robotic Ultrasound System}\label{sec:system}

% \tianpeng{I greatly shortened the system setup section. In other robotic US papers this section is usually rather brief. I think this is good since it gives more weight to the discussion for US scanning pipeline.}

% \subsection{Hardware}

\begin{figure}[htp]
    \centering
    \subfigure[Default State]{\includegraphics[height = 0.40\linewidth]{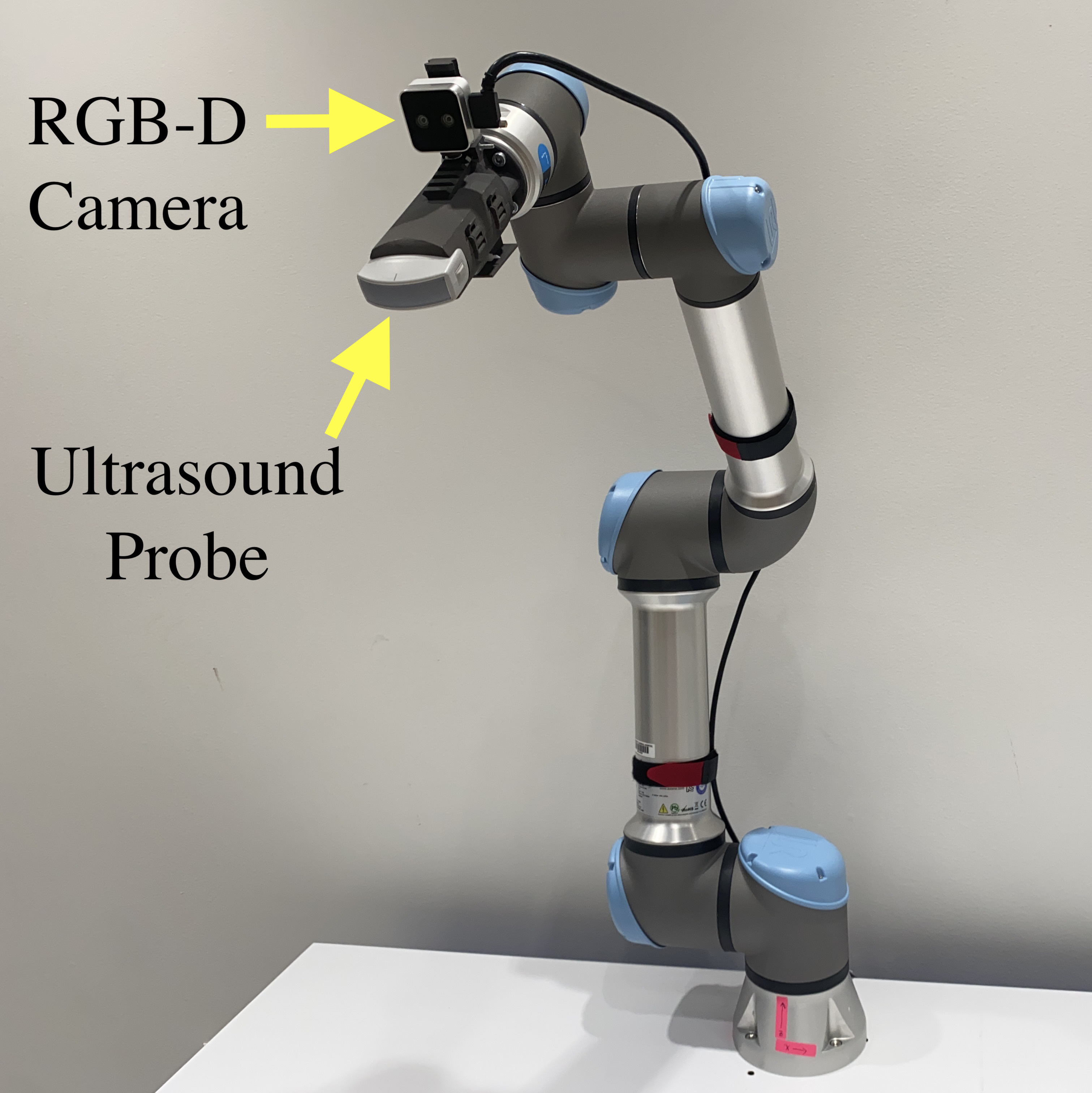} \label{subfig:robot}}
    \subfigure[Contacted State]{\includegraphics[height = 0.40\linewidth]{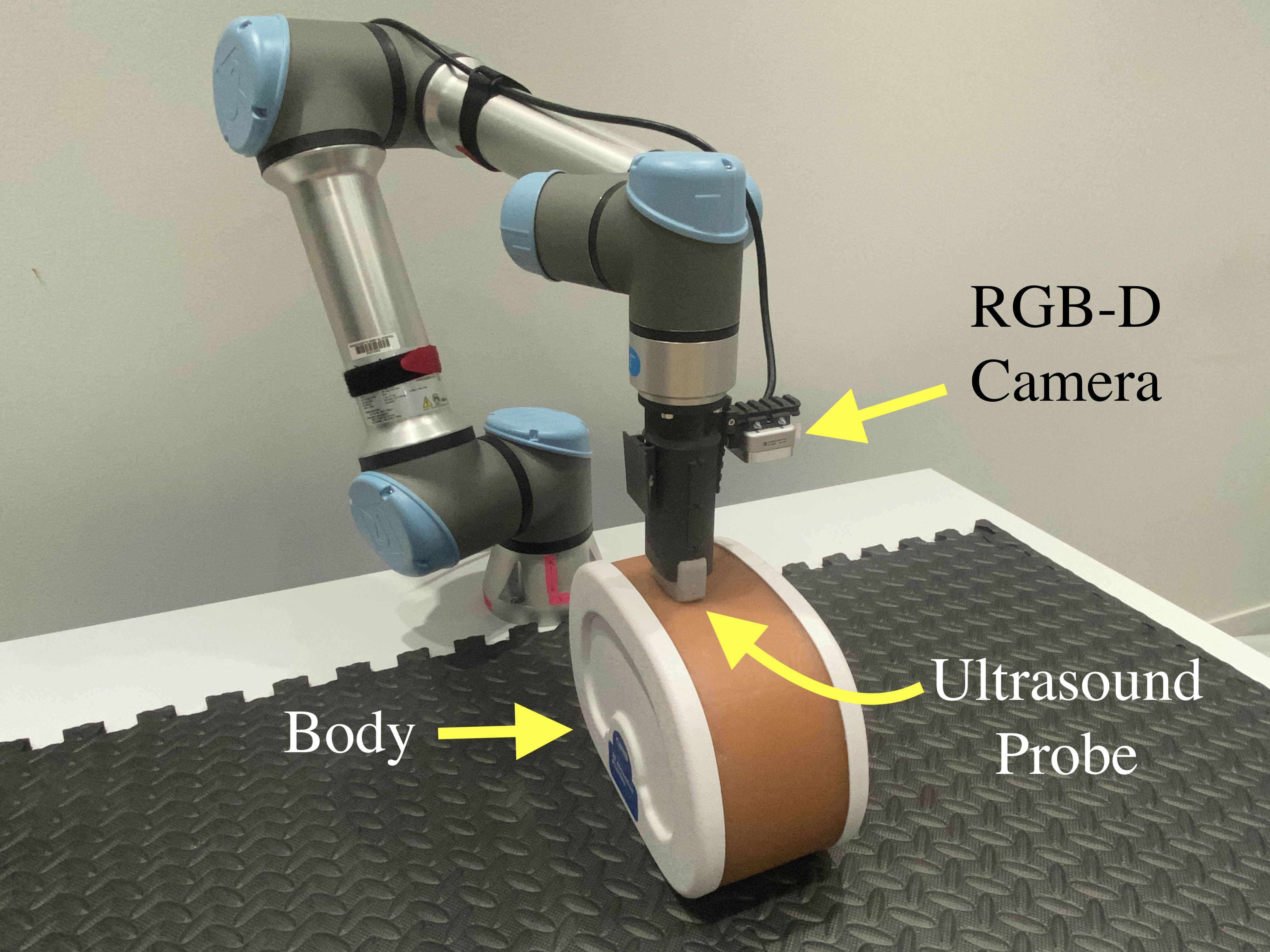}
    \label{subfig:scanning}}
    \caption{Our autonomous ultrasound scanning robot.}
    \label{fig:ultrasound-robot}
\end{figure}

\begin{figure}[htp]
    \centering
        \includegraphics[width = 0.95\linewidth]{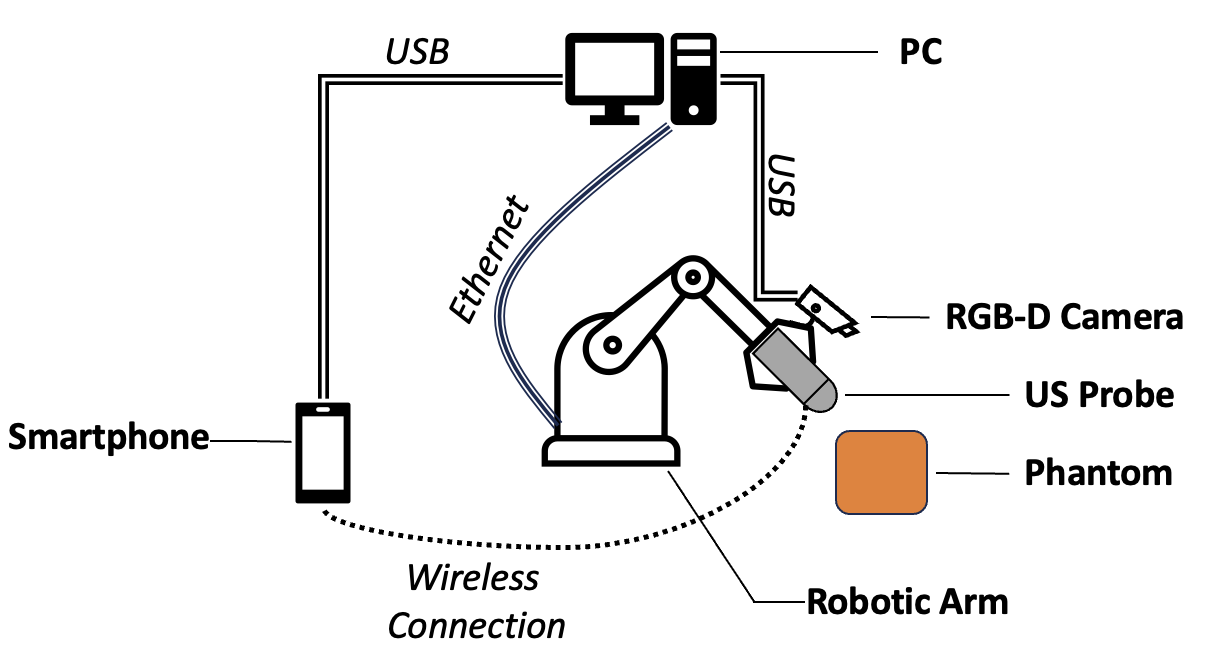}       
    \caption{The illustration describes the interconnectivity of hardware components. Data flow is as follows: RGB-D camera transmits to PC via USB; US probe sends images wirelessly to smartphone, which relays them to PC via USB. PC communicates with the robot arm via Ethernet for state retrieval and control command issuance.}
    \label{fig:system-diagram}
\end{figure}

Our robotic ultrasound scanning system consists of a robotic arm, an ultrasound probe, an RGB-D camera, and a PC that serves as the system controller. Fig.~\ref{fig:ultrasound-robot} illustrates typical configurations of the system and the ultrasound phantom (body). A 3D-printed holder mounted at the robot's end-effector securely encapsulates the ultrasound probe and the RGB-D camera. The holder design ensures the center line of the the US probe's transducer array, the depth axis of the RGB-D camera, and the z-axis of the robot's end-effector, are all aligned. 
The phantom model is CIRS-057A (Computerized Imaging Reference Systems, Inc., Norfolk, VA, USA), a triple modality phantom that can be imaged using CT, MR, and ultrasound. It contains multiple internal features, such as the major organs and blood vessels.  We use the phrases \textbf{\textit{physical body}} or \textbf{\textit{body}} interchangeably with \textbf{\textit{phantom}} in this paper. 

Fig.~\ref{fig:system-diagram} provides a semantic illustration of the data flow between individual hardware components. 
The robot model is the collaborative robotic arm UR5e of Universal Robots (Universal Robots, Odense, Denmark), a 6-joint, 6-DoF robotic arm. The ultrasound probe model is GE Vscan Air CL (GE HealthCare, Chicago, IL, USA), a completely cordless, portable ultrasound probe that can be used for abdominal US scans.  The RGB-D camera model is \textit{Intel\textsuperscript{\textregistered} RealSense\texttrademark~Depth Camera D405} (Intel Corporation, Santa Clara, CA, USA), a stereo depth camera capable of determining the body's 3D location with respect to the robot.

\subsection{Coordinates Systems}

%
% \begin{figure}[htp]
%     \centering
% \includegraphics[height = 0.5\linewidth]{figures/coordinate_systems/anatomical.png}
%     \caption{Anatomical Coordinate System ``Anatomical Planes-en" \textcopyright\ Edoarado(https://commons.wikimedia.org/wiki/File:Anatomical\textunderscore Planes-en.svg), CC BY-SA 3.0. }
%    \label{subfig:anatomical-coordinates}
% \end{figure}

We define the physical (robot) coordinate system to be the same as the medical imaging coordinate. The origin of the physical coordinate system is at the center of the robot's base. The z-axis points vertically upwards, while the x- and y-axes are parallel to the edges of the working surface. 
%See Figure \ref{subfig:full-coordinates} for an illustration.
% The anatomical coordinate system, as illustrated in Figure \ref{subfig:anatomical-coordinates}, defines the standard anatomical positions of a human. Imagine a human patient standing on a level ground. 

The anatomical coordinate system contains three axes characterizing the cardinal directions about the patient: the inferior(feet)-superior(head) axis, the left-right axis, and the anterior(front)-posterior(back) axis. It can be defined by three primary planes.  The \textbf{\textit{axial plane}} is parallel to the ground and separates the patient's inferior from superior. The \textbf{\textit{coronal plane}} is perpendicular to the ground and separates the patient's anterior from posterior. The \textbf{\textit{sagittal plane}} is also perpendicular to the ground and separates the patient's left from right.

The image coordinate system describes how an image is captured in relation to the underlying physical object. In this paper, a medical image $\mathcal{H}$ is characterized by the tuple $(H,\mathbf{s},\mathbf{O}, [\mathbf{a_0}, \mathbf{a_1}, \mathbf{a_2}])$. Here $H$ is a three dimensional array, each element in $H$ represents the content of a rectangular volume in the physical space; $\mathbf{s}=[s_0,s_1,s_2]$ (or $\mathbf{s}=[s_0,s_1]$ for 2D images) is the \textbf{\textit{spacing}} vector, such that $s_i$ represents the physical distance between two adjacent elements(a.k.a. spacing) in  dimension $i$ of $H$. 
%A \textbf{\textit{pixel}} of a 2D image is a location in $H$ specified by an index pair $(i,j)$, and similarly a \textbf{\textit{voxel}} of a 3D image is a location in $H$ specified by the index triple $(i,j,k)$. 
The origin-axes tuple $(\mathbf{O}, [\mathbf{a_0}, \mathbf{a_1}, \mathbf{a_2}])$ specifies the continuous coordinate system for this particular medical image. $\mathbf{O}$ represents the location of pixel $(0,0)$/voxel $(0,0,0)$ of $H$ in the physical space, and $[\mathbf{a_0}, \mathbf{a_1}, \mathbf{a_2}]$ are unit vectors representing the directions of the array axes of $H$ in physical space.

\begin{figure}[htp]
    \centering
\includegraphics[height = 0.5\linewidth]{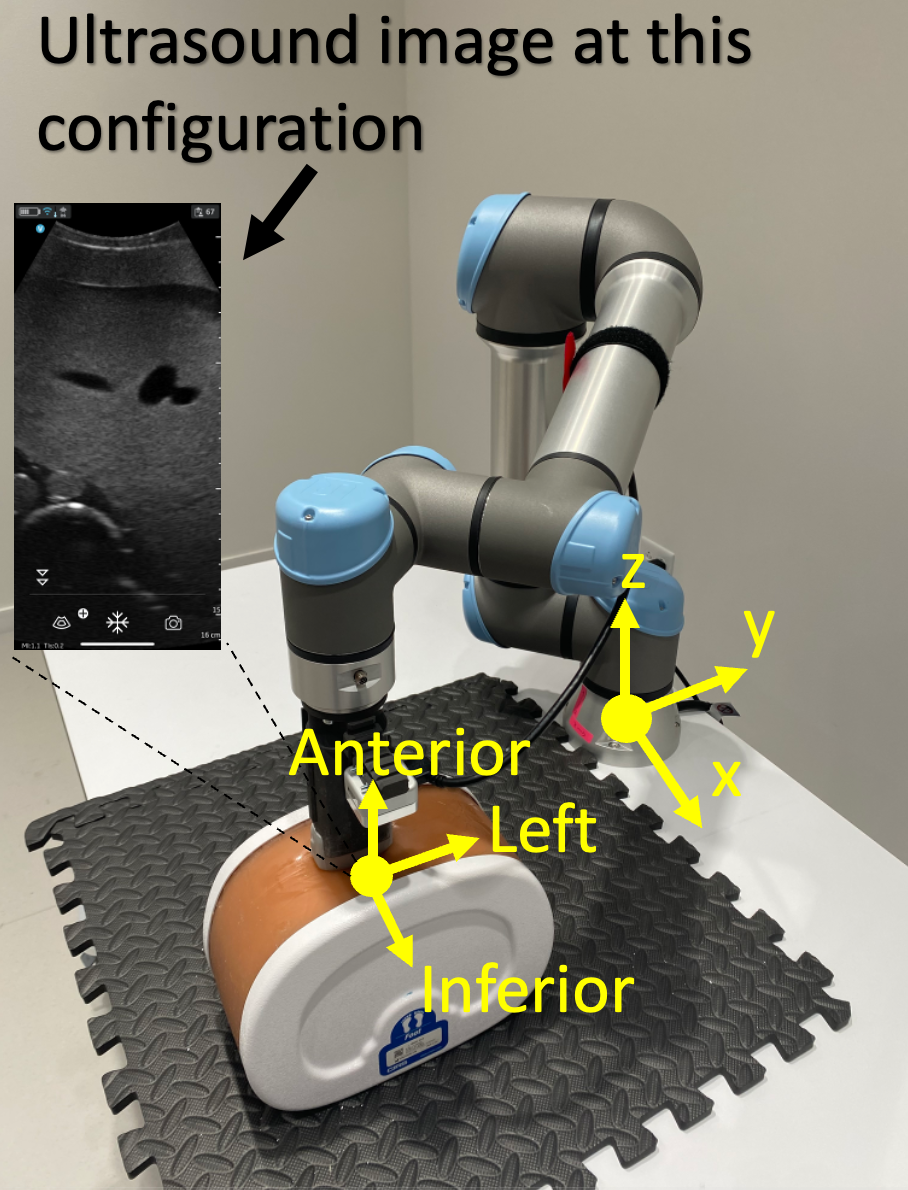}
    \caption{The axis alignment of our system. The axes labeled `x', `y', `z'  with origin at the robot's base constitute our physical coordinate system. The axes labeled `inferior', `left', `anterior' with origin on the body's upper surface constitute the anatomical coordinate system. 
    % The inferior and anterior axes are aligned with the x- and z-axes of the robot base frame. The z-axis of the end-effector is aligned with the negative z-axis of the robot base frame (posterior axis).
    }
    \label{subfig:full-coordinates}
\end{figure}

Figure \ref{subfig:full-coordinates} details the axis alignment of our system during US scanning. The phantom is placed at a fixed location on the working surface as the robot operates, with its inferior and anterior axes aligned with the x- and z-axes of the robot base frame. The z-axis of the robot's end-effector is always aligned with the negative z-axis of the robot base frame (body's posterior axis), such that the ultrasound images are always axial views of the body. This setting assumes the US probe's orientation remains constant during scanning, which is common among phantom studies in RUSS research as it allows for simple low-level motion controls\cite{von2020robotized}. The remainder of the paper focuses on presenting our US imaging pipeline under the system configurations described above.  

% with the image axes aligned with the standard directions of liver ultrasound scans, i.e., the upward direction in the image corresponds to the physical upward direction and the front of the body, while the left/right directions in the image corresponds to the left-/right-hand side of the body.

\section{Autonomous Ultrasound Imaging Pipeline}\label{sec:pipeline}

% \begin{figure}[htp]
%     \centering
%         \includegraphics[width = 0.7\linewidth]{figures/system_diagrams/pipeline_new_v2.png}\label{subfig:pipeline}   
%     \caption{The proposed autonomous US imaging pipeline for liver follow-up scan. The follow-up scan procedure contains three steps: (i) initial contact, (ii) coordinate mapping, and (iii) target imaging. Models are pre-trained.}
%     \label{fig:pipeline}
% \end{figure}

% We propose an autonomous robotic ultrasound system integrated with AI models for effective and high-quality follow-up liver ultrasound scans. This work is a pilot phantom study that lays down the foundation for real-patient applications in the future. 

% \tianpeng{I moved the problem setting directly before introducing the pipeline.}

In this work, we are interested in achieving high-quality autonomous US imaging for follow-up liver patient scan, assuming that:
\begin{itemize}
    \item A CT image of the phantom is available.
    \item The \textbf{\textit{hepatic veins}}(HV) vessels are annotated in the CT image.
    \item Targets to be scanned by US are specified by their 3D coordinates in the CT.
\end{itemize}
Given the above, our goal is to have the robot navigate the US probe to a proper location on the body, and take a sequence of US images containing the target. This procedure must be fully autonomous, without explicit human control on the robot's movement (e.g., through tele-operation). 
% Recall our goal is that, given the target location in CT coordinates, capture a sequence of US images on the physical body about the target. As discussed in Section \ref{sec:problem-setting}, this goal can be broken down into three sub-problems: initial contact, coordinate mapping, and target region scan. 

We develop an autonomous US scanning pipeline, as illustrated in Fig.~\ref{fig:big-system-semantics}, that achieves our goal. We train segmentation models that identify the HV structure in US images and use them in the liver follow-up scans. The follow-up scan consists of three stages: 
\begin{enumerate}
    \item Make initial contact with the body.
    \item Compute the coordinate mapping between CT and physical frame.
    \item Perform target region scan.
\end{enumerate}

In particular, the second stage is the most important component in the follow-up scan, where we employ AI technique to automatically acquire a 3D model of HV and compute the CT-physical mapping through image registration. The rest of this section is organized as follows:

\begin{itemize}
    \item Section \ref{subsec:vessel-segmentation} describes the training of deep segmentation networks for HV prior to the execution of follow-up scan. 
    \item Section \ref{subsec:initial-contact} describes the initial contact method using RGB-D camera view. 
    \item Sections \ref{subsec:HV-localization} to \ref{subsec:coordinate-mapping} are dedicated to solving the coordinate mapping problem. 
    \begin{itemize}
        \item Section \ref{subsec:HV-localization} describes the HV search procedure which localizes HV in the body.
        \item Section \ref{subsec:HV-acquisition} describes how we construct a 3D HV volume from 2D HV segmentation masks of US images.
        \item Section \ref{subsec:coordinate-mapping} describes the computation of the transformation $^{p}T_{c}$ between the CT coordinates and physical coordinates by registering the 3D HV volume construsted from US images with the ground-truth HV annotation in CT.
    \end{itemize}   
    \item Section \ref{subsec:target-localiation-imaging} discusses target localization and imaging given the coordinate mapping. 
\end{itemize}

\subsection{HV Vessel Segmentation}\label{subsec:vessel-segmentation} 

\begin{figure}[htp]
    \centering
    %    \subfigure[Liver Anatomy]{ \includegraphics[width = 0.6\linewidth]{figures/liver_anatomy_v2.jpeg}\label{subfig:anatomy}}
        \subfigure[Target location in CT]{
        \includegraphics[height = 0.5\linewidth]{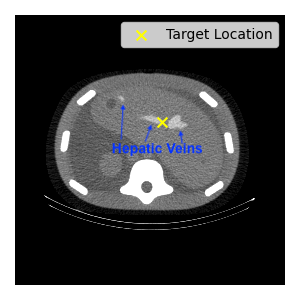}\label{subfig:example_target_loc}
    }
    \subfigure[Ultrasound image]{
        \includegraphics[height = 0.5\linewidth]{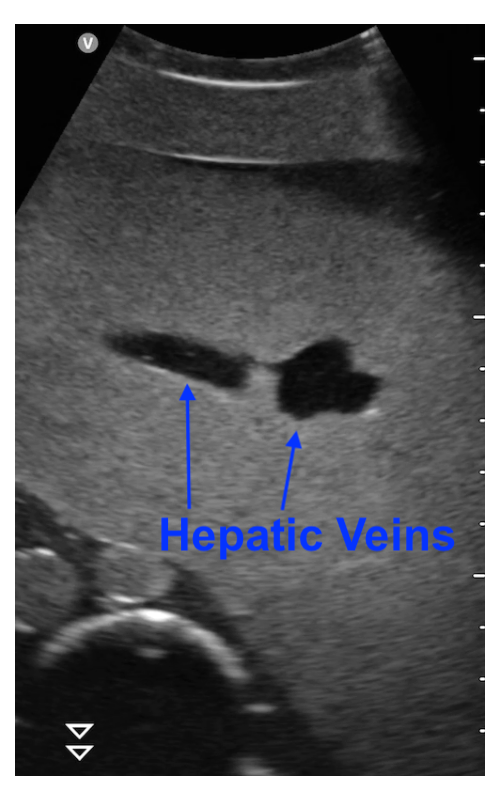}\label{subfig:example_us_scan}
    }
    \caption{
    % (a) ``Cenveo - Drawing Liver anatomy and vascularisation - English labels" \textcopyright\  Cenveo(https://anatomytool.org/content/cenveo-drawing-liver-anatomy-and-vascularisation-english-labels), CC BY 4.0. Image is modified to highlight the branching point of upper hepatic veins. 
    (a) A target location in the CT image near the branching point. (b) Ultrasound image taken by the robot that matches the target in (a). 
    }
    \label{fig:liver-anatomy}
\end{figure}
The HV is arguably the most important image feature in our US imaging pipeline. We use HV in liver ultrasound images as the key landmark to navigate the US probe on the body, and most importantly, determine the the CT-physical coordinate mapping $^{p}T_{c}$. The HV is a group of major blood vessels in liver that originate from multiple liver sections and ultimately join the inferior vena cava. These vessels are easily distinguishable in various perspectives of CT and ultrasound images. In particular, the first branching point of the upper hepatic veins exhibits an iconic `bunny ear' shape in axial ultrasound images as shown in Fig.~\ref{subfig:example_us_scan}. Literature on US-CT registration has demonstrated great success using HV as the feature for registration\cite{he2023robust,haque2016automated}.
\begin{figure}[t]
    \centering
    \subfigure[Annotation for full-vessel model]{
        \includegraphics[width = 0.75\linewidth]{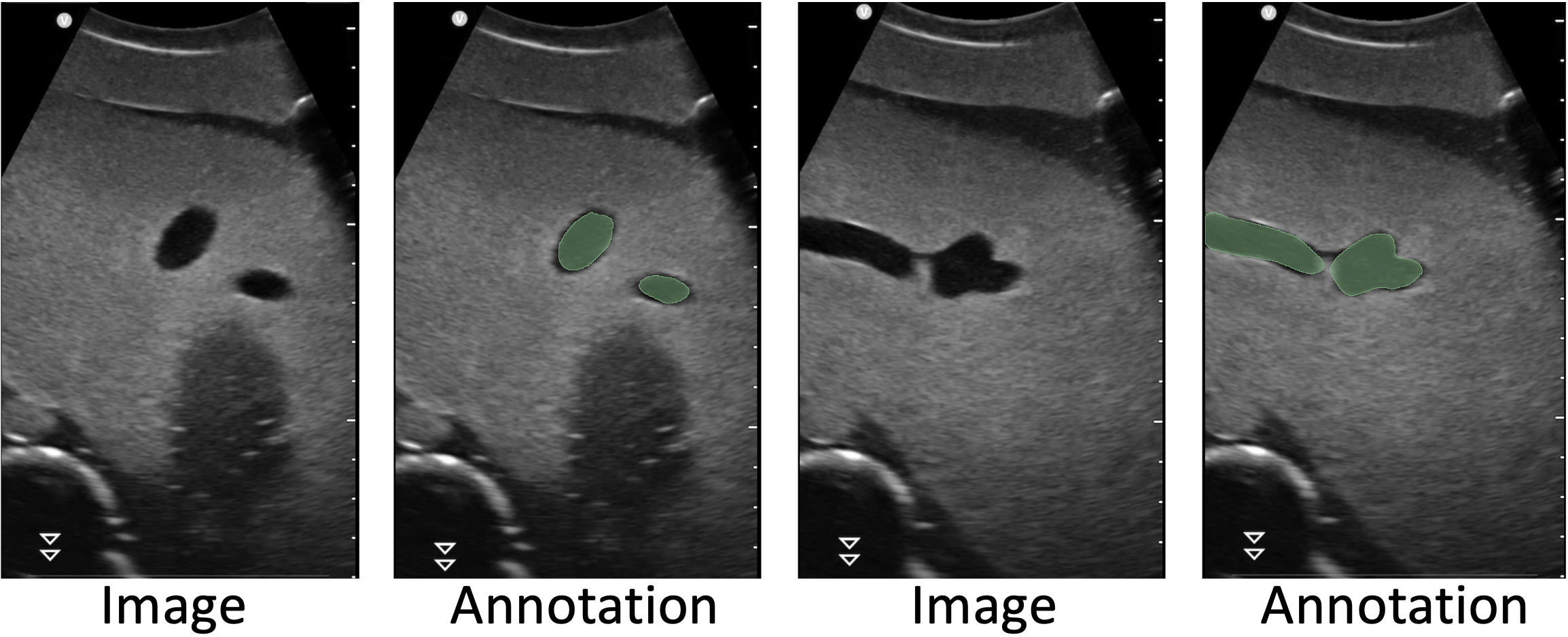}\label{subfig:full-annoation}
    }
    \subfigure[Annotation for branching point model]{
        \includegraphics[width = 0.75\linewidth]{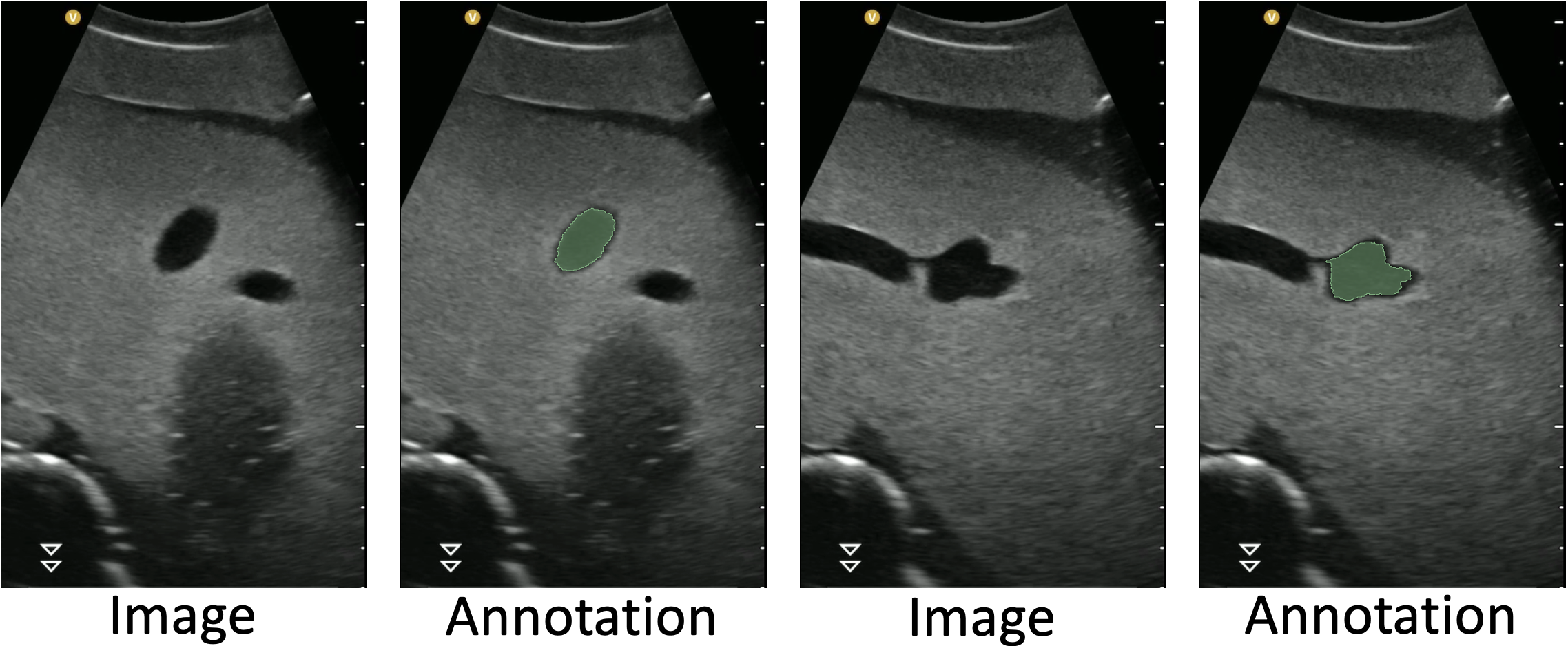}\label{subfig:key-annoation}
    }    
    \caption{The ground-truth annotations for our segmentation models (overlaid in green). 
    % The subfigure for branching point label is generated using Dropbox/linear scan data/25x25/
    }\label{fig:manual-label}
\end{figure}

% \tianpeng{This section needs more specification of training details.}

We train two segmentation networks with the U-Net architecture\cite{ronneberger2015u} to identify HV in axial-view US images. One is the \textbf{\textit{full-vessel model}}, which segments the entire system of HV vessels. The full-vessel model is needed for 3D HV model acquisition and final target imaging(Sections \ref{subsec:HV-acquisition} and \ref{subsec:target-localiation-imaging}). The other is the \textbf{\textit{branching point model}}, which segments only the HV vessels along the middle hepatic vein, near the first branching point of upper hepatic veins. The branching point model is only used to find and centralize the HV branching point in the US probe's field of view(Section \ref{subsec:HV-localization}). Figure \ref{fig:manual-label} illustrates the expected segmentation from these two models.  

We use the robot to acquire a set of axial-view US images sampled from a uniform 50x50 grid on the phantom's surface as the training data. We then manually annotate the images to create labels for the  branching point segmentation and full-vessel segmentation (Fig. \ref{fig:manual-label}). 
% We build training set a variety of acquisition qualities along probe pressure on surface to assure the robotic scan probe would have low image qualities.
% We show the images with different pressure on surface in Fig. x. Total of xx images and labeled data used for training set and xx data for validation set. 
% \tianpeng{I forgot if we used images with different pressures when training the neural networks?} 
We also employ augmentation techniques including flipping, rotation, and additive Gaussian noise, to diversify the training data and enable the models to learn features that are invariant to contact force, noise levels, and probe rotation in the axial plane. 
We train the networks each for 1000 epochs with a batch size of 32 using Adam optimizer. The training loss is cross entropy as defined in\cite{ronneberger2015u}. 
The data collection and training for the segmentation networks are carried out prior to the actual execution of follow-up scans.

\subsection{Initial contact}\label{subsec:initial-contact}

\begin{figure}[t]
    \centering
    % \subfigure{
    %     \includegraphics[width = 0.4\linewidth]{figures/body_seg_config.JPG}
    % }
    \subfigure{
        \includegraphics[width = 0.95\linewidth]{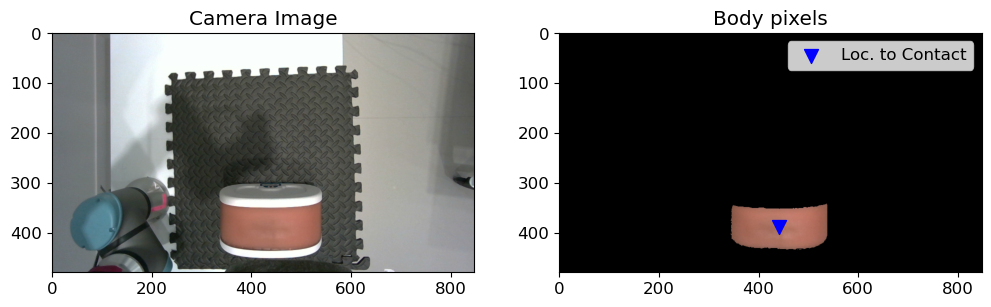}
    }
    \caption{An example of body pixel segmentation when determining the location for initial contact. The location to make contact with, a.k.a. the centroid of body pixels, is indicated by the blue `$\blacktriangledown$' in the right subfigure. 
    % The top figure shows the configuration at which the camera image is taken. % The top figure is captured by phone camera. The bottom figures are captured by RGB-D camera in lab.
    }
    \label{fig:body-seg}
\end{figure}
The first step of the follow-up scan procedure is to have the robot make contact with the body and apply a desirable amount of contact force. Using the sensory output from the RGB-D camera, the system localizes and makes initial contact with the body through the following steps.

\begin{enumerate}
    \item The robot moves to the default pose and orient the end-effector to face the camera vertically downwards.
    \item By thresholding the pixel values, we segment out the pixels corresponding to the body in the color image seen by the camera. See Fig.~\ref{fig:body-seg} for an illustration. The thresholds on the pixel values are designed to select pixels within the orange-brown spectrum.  We then define the centroid of these pixels as the \textbf{location to contact} for the probe. 
    % Appendix \ref{append:body-segmentation} provides a detailed description of the thresholding method.
    \item  
    We use the point cloud computation function from \textit{Intel\textsuperscript{\textregistered} RealSense\texttrademark~SDK 2.0} to compute the 3D coordinates of the \textbf{location to contact} with respect to the camera, and then transform the coordinates to be represented in the robot base frame.% \footnote{Prior to the experiments, we manually measure the relative offset between the RGB-D camera and the robot end-effector and record it as a constant the configuration files. And since the \texttt{ur\_rtde} toolbox allows us to read the relative pose between the robot end-effector and base,  we effectively knows the transformation between the camera coordinate frame and the robot base coordinate frame at any time.}
    \item The robot moves the probe to hover above the \textbf{location to contact} inferred from step 2), then drives the probe vertically downwards until a contact is detected by the sensors in the end-effector.
    \item After contact, the robot slowly presses the probe into the body until the contact force reaches a desired level, which is typically 20N in our experiments.
\end{enumerate}
\begin{remark}
The color thresholding method in step 2) works well in our experiments since the phantom is the only orange-brown colored object visible to the camera, but this method is admittedly \textit{ad-hoc} to the experimental conditions in this work. 
When extending our method to human patients, 
% the color thresholding should be replaced by human pose estimation or other more robust methods to identify the body within the camera's field of view. 
% In real-human scans,
the location to contact could be inferred using human pose estimation methods. For instance, methods like \cite{keller2023skin} can be applied to estimate the patient's skeleton pose from the camera view, then the location to contact can be selected with respect to the patient's skeleton frame. These alternative initial contact solutions are left for the future work. 
\end{remark}
\begin{remark}
Extensive lab observation indicates that exerting a vertically downward initial contact force of 20N on the phantom is sufficient for great ultrasound image quality in our experiments. Though on real patients, more complex force control is necessary to ensure the patients' comfort while preserving ultrasound imaging quality\cite{finocchi2017co,ning2021autonomic}. Such force control method is left for the future work.  
\end{remark}

\subsection{HV Search}\label{subsec:HV-localization}
\begin{figure}[t]
    \centering
    % \subfigure[HV search waypoints]{
    %     \includegraphics[width = 0.3\linewidth]{figures/hv_search/hv_search_waypoints.png}\label{subfig:hv-search-waypoints}
    % }
    \includegraphics[width = 0.6\linewidth]{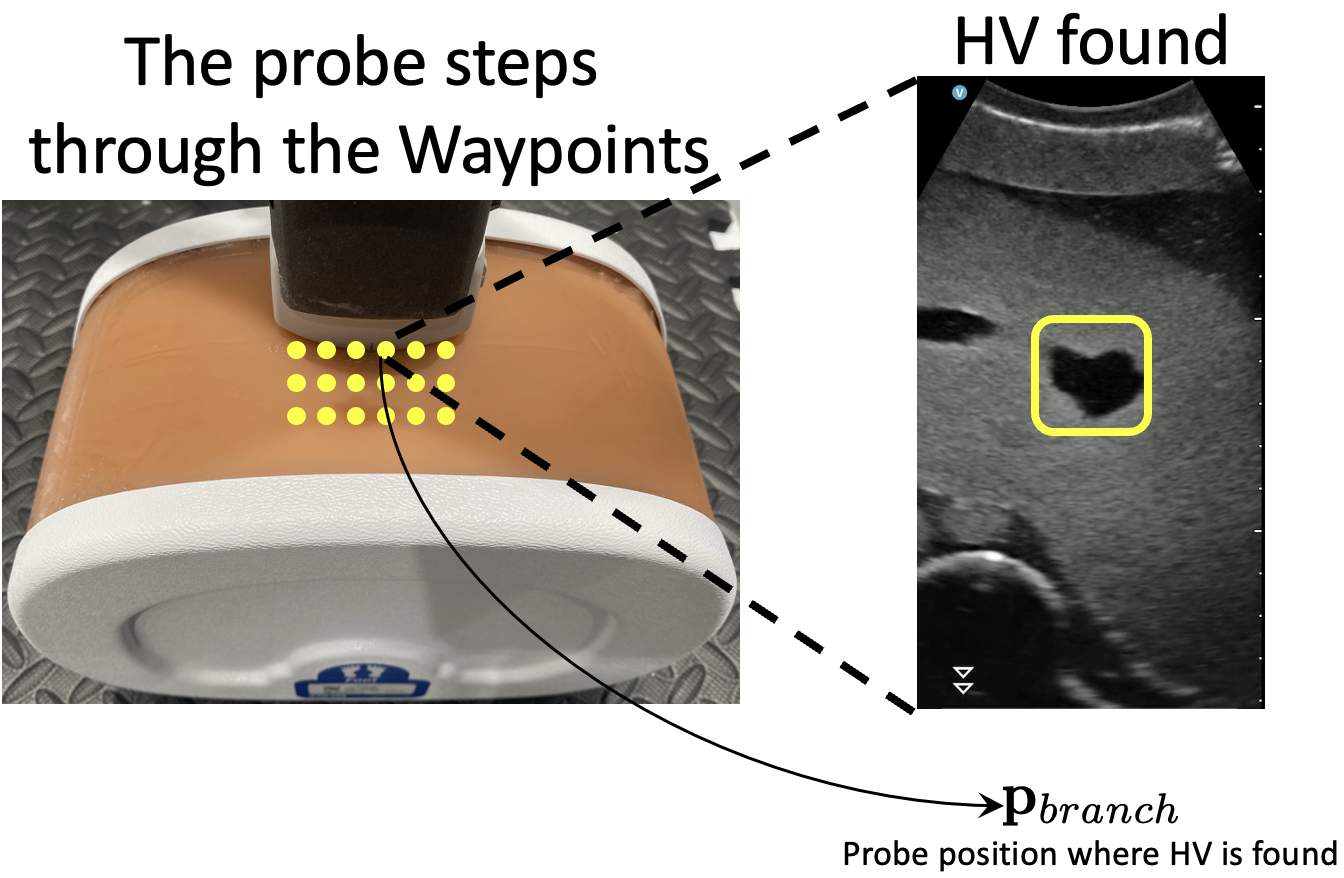}\label{subfig:hv-found}
    \caption{Illustration of the HV search algorithm. 
    % The location $\mathbf{p}_0$ marked by the blue `$\blacktriangledown$' corresponds to the contact location in Fig. \ref{fig:body-seg}.
    The robot moves the probe to step through a set of waypoints in the neighborhood of $\mathbf{p}_0$ (the contact location in Fig. \ref{fig:body-seg}) and uses the branching point model to determine whether the HV branching point is found in the US image. Upon the detection of HV branching point, the robot centralizes the branching point pixels in the US probe's field of view, then returns the probe location $\mathbf{p}_{branch}$.
    }
    \label{fig:hv-search}
\end{figure}
\begin{algorithm}[t]
    \caption{Hepatic Vein Search and Centralization}\label{alg:hv_search}
    \Required{Branching point model $\mathcal{M}_{b}$; initial probe location $\mathbf{p}_0$; HV detection threshold $\epsilon_0$; step size $d$; HV centralization threshold $\epsilon_1$; ultrasound image shape $(l_x,l_y)$.
    }
    \KwOut{Probe location $\mathbf{p}_{branch}$ upon success, or \textbf{Failure}.}
    \LinesNumbered

    Generate waypoints $\mathcal{W}$ centered at $\mathbf{p}_0$.

    $hv\_found\gets False$
    
    \ForEach(\algcomments{Search for HV}){
        $\mathbf{w} \in \mathcal{W}$
    }{
        Move the probe to $\mathbf{w}$ and take ultrasound image $img$. Segmentation mask $h\gets \mathcal{M}_{b}(img)$.

        Erase all non-zero elements in $h$ except for the largest connected component.
        
        $A\gets \sum_{i=1}^{l_x} \sum_{j=1}^{l_y} h_{ij}$, the area the largest connected component.

        \If{$A\geq \epsilon_0$}{

            $hv\_found\gets True$
            
            \Break
        }
    }
 
    \If{$hv\_found = False$}{

        Return \textbf{Failure}.
    }

    \Repeat(
    \algcomments{Centralize HV}
    ){$|c_x-l_x/2|\leq\epsilon_1$}{

        Take ultrasound image $img$. $h \gets \mathcal{M}_{b}(img)$
 
        % $\{(x_1,y_1),(x_2,y_2),...(x_m,y_m)\}\gets \{(i,j)| h_{ij}=1\}$, the pixels for HV in the $img$.

        $(c_x,c_y)\gets$ The center non-zero pixels in $h$.
        
        \If{$c_x\geq l_x/2$}{Moves the probe to the right for distance $d$.}\Else{Moves the probe to the left for distance $d$.}
    }

    return current probe location $\mathbf{p}_{branch}$
\end{algorithm}

After making contact with the body, the robot carries out a search algorithm to localize the first branching point of upper hepatic veins in the phantom. The search method is illustrated in Figure \ref{fig:hv-search} and described in Algorithm \ref{alg:hv_search}. Upon successful execution of the algorithm, the ultrasound probe would eventually aim at the middle-hepatic vein near the branching point, creating an ideal condition for the subsequent HV acquisition.

Algorithm \ref{alg:hv_search} starts by generating a set of waypoints centered around $\mathbf{p}_0$, the probe location after initial contact (line 1). The robot then positions the probe to step through these waypoints. At each of the waypoints, we acquire a US image and compute the predicted segmentation mask for HV in the image by the branching point model (line 4). The segmentation mask is a binary 2D array with the same shape as the US image, in which the pixels corresponding to the HV have value 1, and otherwise 0. As the predicted masks sometimes contain small, erroneous clusters of pixels that do not belong to HV, we denoise the segmentation by erasing all non-zero pixels in the mask except for the largest connected component (line 5). The area of the largest connected component, i.e., the number of pixels in the component, is then compared with a pre-specified threshold $\epsilon_0$ to determine whether the HV is found or not (line 6-9). We usually set $\epsilon_0=4000$ under our experiment conditions, where the shape of US images is $(1080,500)$.

If the HV is not found after exhausting all the waypoints, the algorithm returns \textbf{Failure} (line 10-11). Otherwise, the robot adjusts the probe location along the left-right anatomical axis through a simple position feedback control algorithm, until the HV is centralized in the ultrasound image (line 12-19), then returns the probe location $\mathbf{p}_{branch}$ upon success (line 20).

%%%% Add the following remark if the reviewers asked. %%%%
% \begin{remark}
% Using the branching point model for HV segmentation in Algorithm \ref{alg:hv_search} is crucial since it ensures the probe is aimed at the middle-hepatic vein at the start of the subsequent HV acquisition step (Section \ref{subsec:HV-acquisition}), where we hope to construct a 3D model for HV that captures all of left-, middle-, and right-hepatic veins using the full-vessel model. In contrast, if the branching point model is replaced by the full-vessel model, we observed that the final probe position often biases heavily towards the left-hepatic branch, which produces an 3D IVC model that captures the left-hepatic branch only in HV acquisition and the resulting coordinate mapping (Section \ref{subsec:coordinate-mapping}) accuracy is poor. 
% \end{remark}

\begin{remark}
We experimented with various simple waypoint generation techniques, such as evenly spaced waypoints along the superior-inferior direction and regular grids, and all seem to be fairly robust in finding the HV. In fact, the HV is often already visible in the ultrasound image immediately after the initial contact, since it is rather close to the center of the phantom's surface. Nonetheless, locating the HV on real patients is more challenging as the geometry and composition of real human bodies are much more complex than the phantom, and consequently the search algorithm needs to be crafted more carefully.
\end{remark}

\subsection{HV Acquisition}\label{subsec:HV-acquisition}

\begin{figure}[t]
    \centering
       \subfigure[Acquisition Waypoints]{
        \includegraphics[width = 0.45\linewidth]{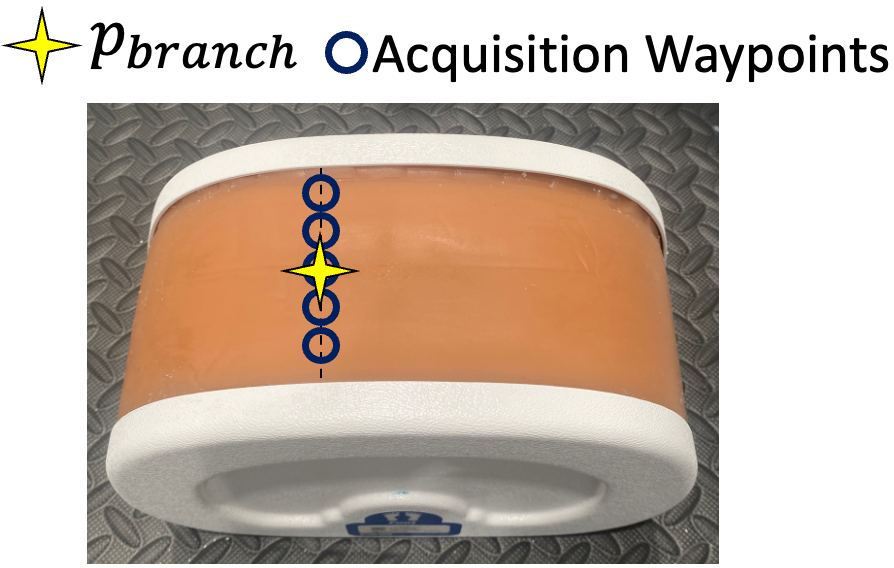}\label{subfig:hv-acquisition-waypoints}
    }
       \subfigure[Stacking Segmentation Masks]{
        \includegraphics[width = 0.45\linewidth]{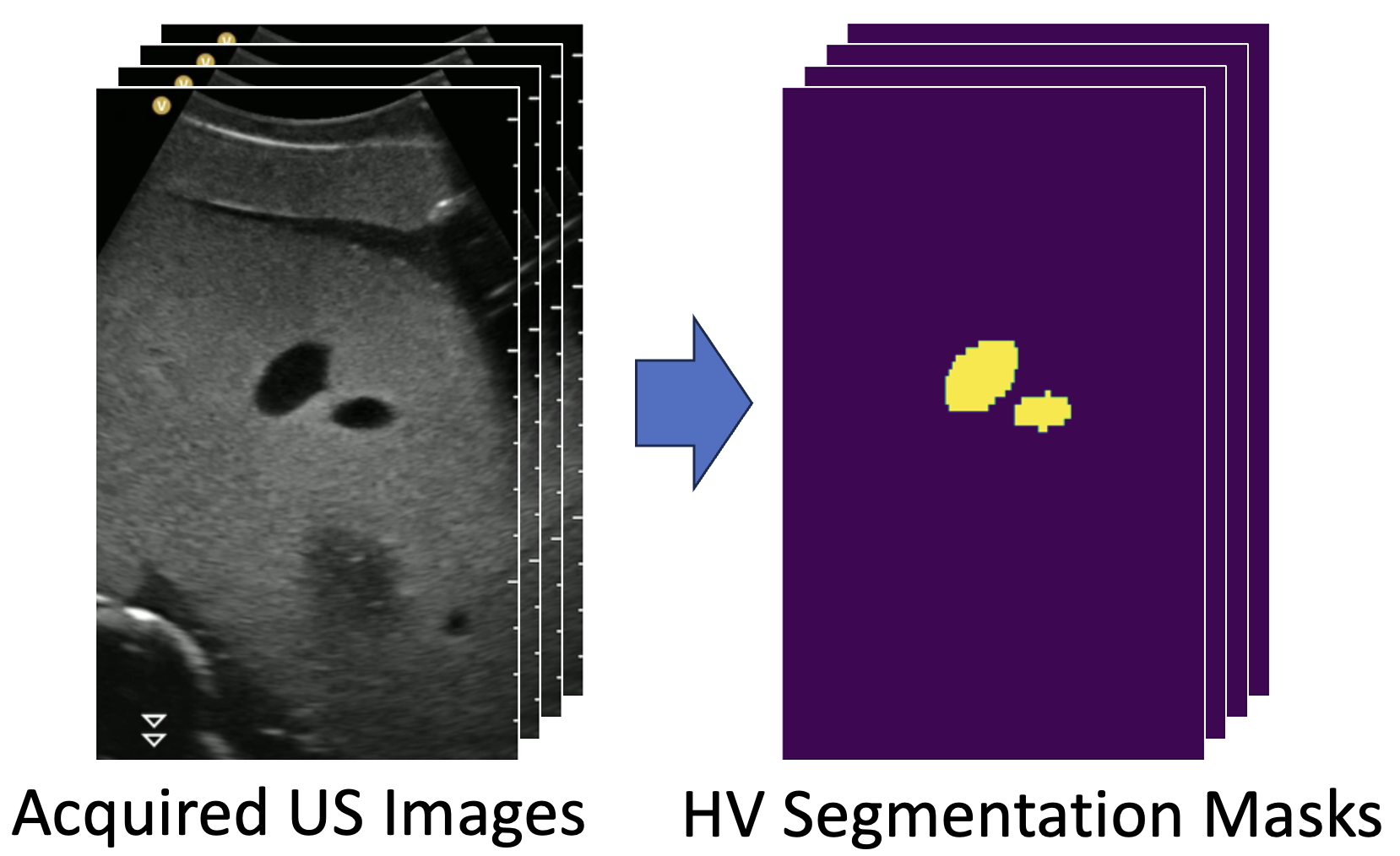}\label{subfig:hv-acquisition-stack}
    }
    
    \subfigure[HV acquisition result($\mathcal{H}_{US}$)]{
        \includegraphics[width = 0.4\linewidth]{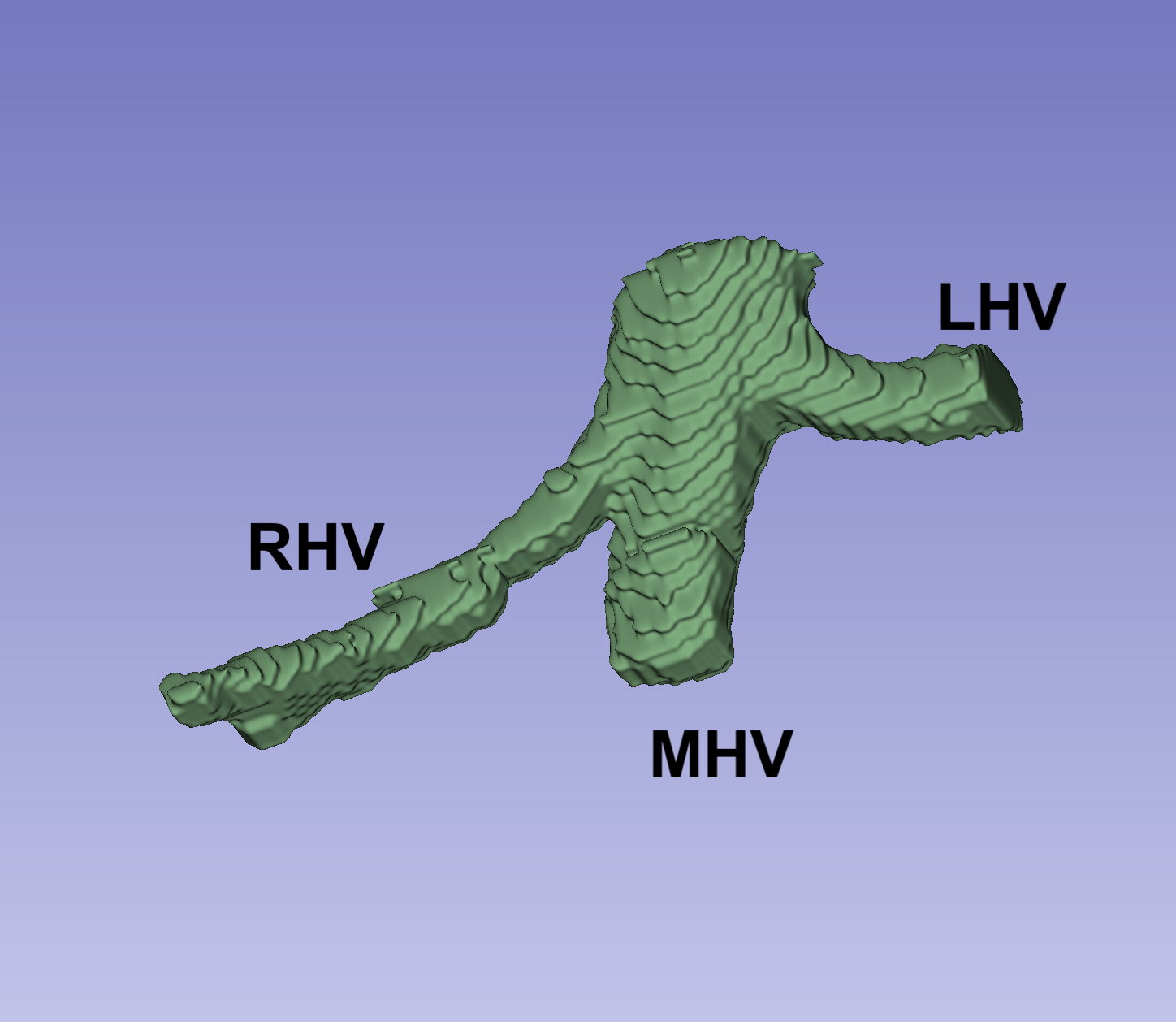}\label{subfig:hv-acquisition-result}
    }
    \subfigure[HV Ground Truth($\mathcal{H}_{CT}$)]{
        \includegraphics[width = 0.4\linewidth]{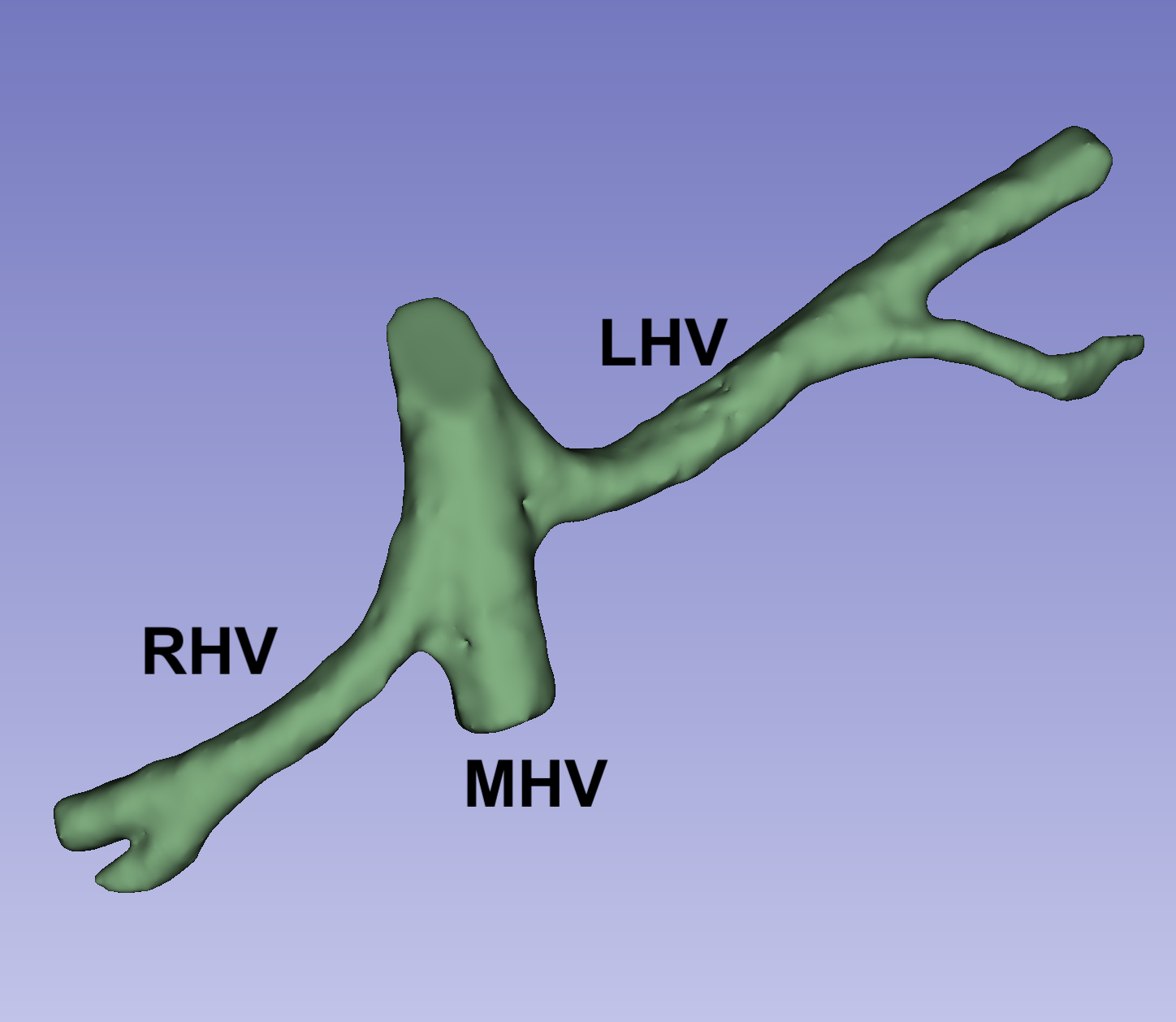}\label{subfig:hv-CT-groundtruth}
    }
    \caption{HV acquisition. The robot aqcquires a sequence of US images at waypoints centered around $\mathbf{p}_{branch}$ along the superior-inferior direction. The HV segmentation masks of the acquired images are stacked together to create the 3D HV model in (c). Subfigure (d) shows the ground truth HV structure in the CT image for comparison. The meaning of labels: \textbf{RHV}$\rightarrow$right-hepatic vein, \textbf{MHV}$\rightarrow$middle-hepatic vein, \textbf{LHV}$\rightarrow$left-hepatic vein. 
    % These figures are generated using data_Jan7_0
    }
    \label{fig:hv-aquisition}
\end{figure}

\begin{algorithm}[t]
    \caption{HV Acquisition}\label{alg:hv_acquisition}
    \Required{Full-vessel segmentation model $\mathcal{M}_{F}$; initial probe location $\mathbf{p}_{branch}$; number of samples to collect $n$; acquisition distance $L$; ultrasound image shape $(l_x,l_y)$; the spacing vector of a (2D)ultrasound image $\mathbf{v}=[v_x,v_y]$ provided by ultrasound probe manufacturer; $[\mathbf{\hat{x}},\mathbf{\hat{y}},\mathbf{\hat{z}}]$, unit vectors representing the axes for the physical coordinate system.
    }
    
    \KwOut{$\mathcal{H}_{US}=(H,\mathbf{s},\mathbf{O},[\mathbf{a_0}, \mathbf{a_1}, \mathbf{a_2}])$, a binary 3D image of HV from US;}
    \LinesNumbered

    Generate a set of equally-spaced waypoints $\mathbf{w_0},\mathbf{w_1},...,\mathbf{w_{n-1}}$ along the inferior-superior axis   centered at $\mathbf{p}_{branch}$, satisfying $||\mathbf{w_0}-\mathbf{w_{n-1}}|| = L$.
    
    \For(\algcomments{Capture images of HV}){
        $i=0,1,2,...,n-1$
    }{

        The robot moves the probe to $\mathbf{w_i}$ and take ultrasound image $img$. 
        
        Compute the binary segmentation mask: $h\gets \mathcal{M}_{F}(img)$.

        $H_{ijk}\gets h_{jk}$ for all $1\leq j\leq l_x, 1\leq k\leq l_y$. \algcomments{Saving $h$ as the i'th slice of $H$}
    }

    $\mathbf{s}\gets [L/n, v_x,v_y]$

    $[\mathbf{a_0},\mathbf{a_1},\mathbf{a_2}]\gets[\mathbf{\hat{x}},\mathbf{\hat{y}},-\mathbf{\hat{z}}]$

    $\mathbf{O} \gets \mathbf{w_0} - \frac{l_xv_x}{2}\mathbf{\hat{y}}$
    
    return $\mathcal{H}_{US} = (H,\mathbf{s},\mathbf{O},[\mathbf{a_0}, \mathbf{a_1}, \mathbf{a_2}])$
\end{algorithm}

After locating and centralizing the HV in the ultrasound field of view, the robot captures a sequence of ultrasound images about the HV to construct a 3D image $\mathcal{H}_{US}=(H,\mathbf{s},\mathbf{O},[\mathbf{a_0}, \mathbf{a_1}, \mathbf{a_2}])$ of the HV in the physical space. We call this procedure \textbf{\textit{HV acquisition}}, which is described in Algorithm \ref{alg:hv_acquisition} and illustrated in Fig.~\ref{fig:hv-aquisition}. In particular, Fig.~\ref{subfig:hv-acquisition-result} displays an example of the acquisition result by Algorithm \ref{alg:hv_acquisition}.

The first step of the Algorithm \ref{alg:hv_acquisition} is to generate a sequence of acquisition positions $\mathbf{w_0},\mathbf{w_1},...,\mathbf{w_{n-1}}$ centered at $\mathbf{p}_{branch}$--the probe location reached in the previous step (line 1). As illustrated in Fig.~\ref{subfig:hv-acquisition-waypoints}, the waypoints are placed linearly along the inferior-superior axis, with equal spacing between them. Using the full-vessel model, we predict the HV segmentation masks for the images taken at each of the waypoints (line 3-4), 
% Each of these masks is a 2D binary array, with non-zero array elements indicating HV pixels, and zero array elements indicating non-HV pixels. 
then stack the masks together to create $H$--a 3D binary array capturing the HV structure (line 5). After that, we calculate the coordinate parameters for the 3D image (line 6-8). The $s_0$ element of the spacing vector $\mathbf{s}$ is the spacing between the waypoints, while the $s_1,s_2$ elements are the same as the spacing values of ultrasound images provided by the ultrasound probe manufacturer (line 6). The vectors $[\mathbf{a_0},\mathbf{a_1},\mathbf{a_2}]$ represent the axes of the image coordinate frame for $\mathcal{H}_{US}$, which are determined according to our experiment setting described in Section \ref{sec:system}.
Since the center of the ultrasound transducer array always aligns with the center line of the ultrasound images, the physical location $\mathbf{O}$ of the origin of $H$ would be half the image width to the left of $\mathbf{w_0}$ (line 8). Finally, the algorithm composes the image array $H$ with the coordinate parameters to become $\mathcal{H}_{US}$ and returns (line 9).

Figs.~\ref{subfig:hv-acquisition-result} and \ref{subfig:hv-CT-groundtruth} compare a typical acquired HV model built by Algorithm \ref{alg:hv_acquisition} with the ground truth HV annotation in CT. Notice that, due to the limited field of view of our ultrasound probe, the acquired 3D image $\mathcal{H}_{US}$ only covers a portion of the ground truth $\mathcal{H}_{CT}$. Nevertheless, $\mathcal{H}_{US}$ does capture the most significant structures near the branching point, including all three branches of the upper hepatic veins.

\subsection{3D Vessel Registration and Coordinate Mapping Computation}\label{subsec:coordinate-mapping}

\begin{figure}[t]
    \centering
    \subfigure[Initial Images]{
        \includegraphics[width = 0.295\linewidth]{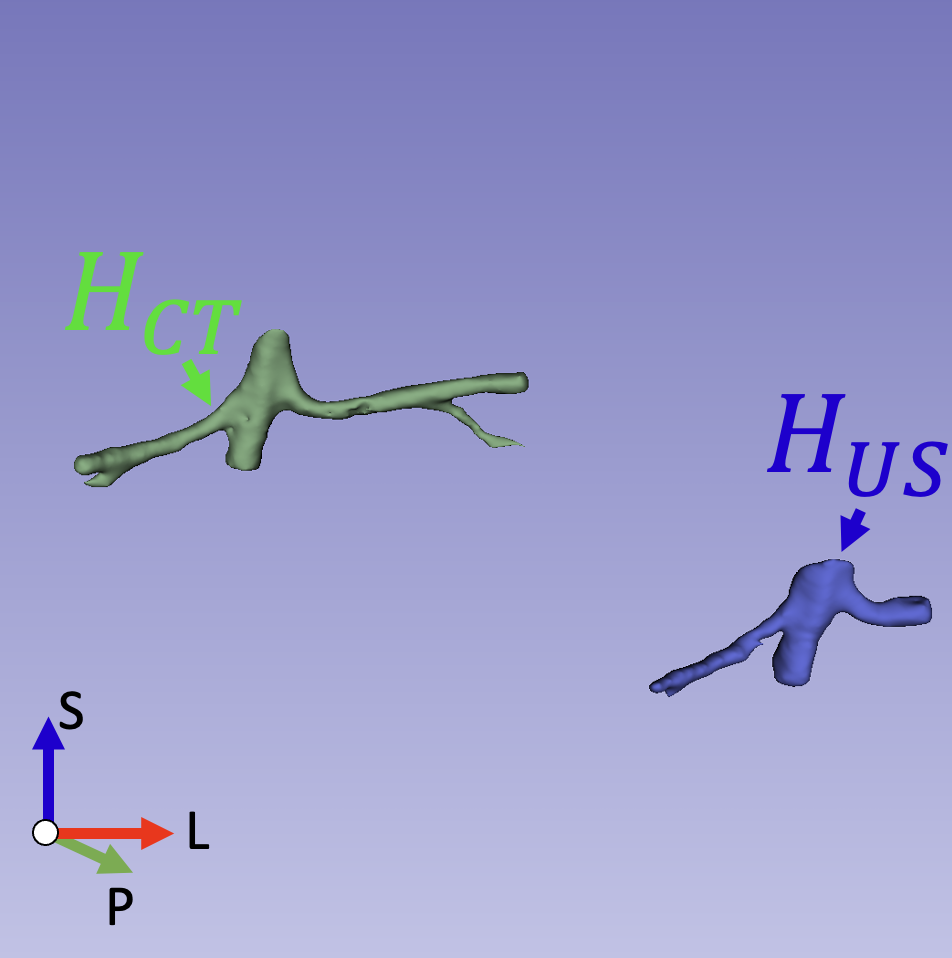}\label{subfig:pre-shift}
    }
    \subfigure[After Translation]{
        \includegraphics[width = 0.295\linewidth]{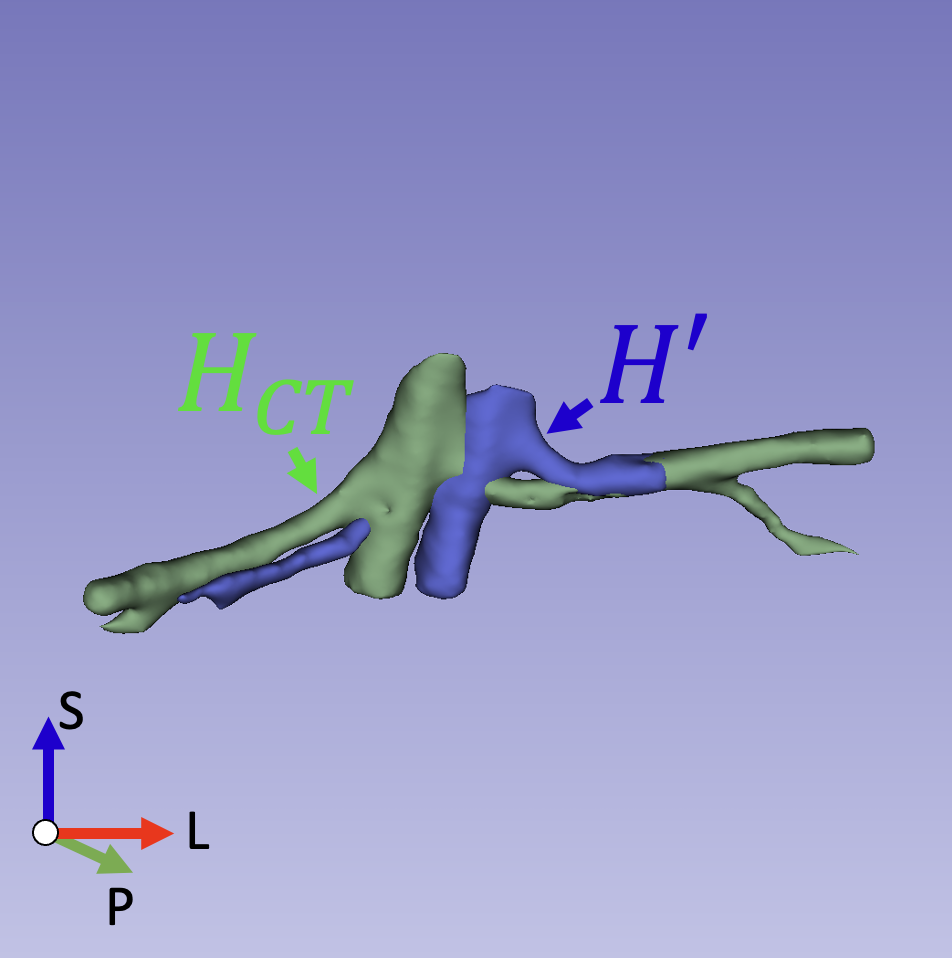}\label{subfig:pre-registration}
    }
    \subfigure[After Registration]{
        \includegraphics[width = 0.295\linewidth]{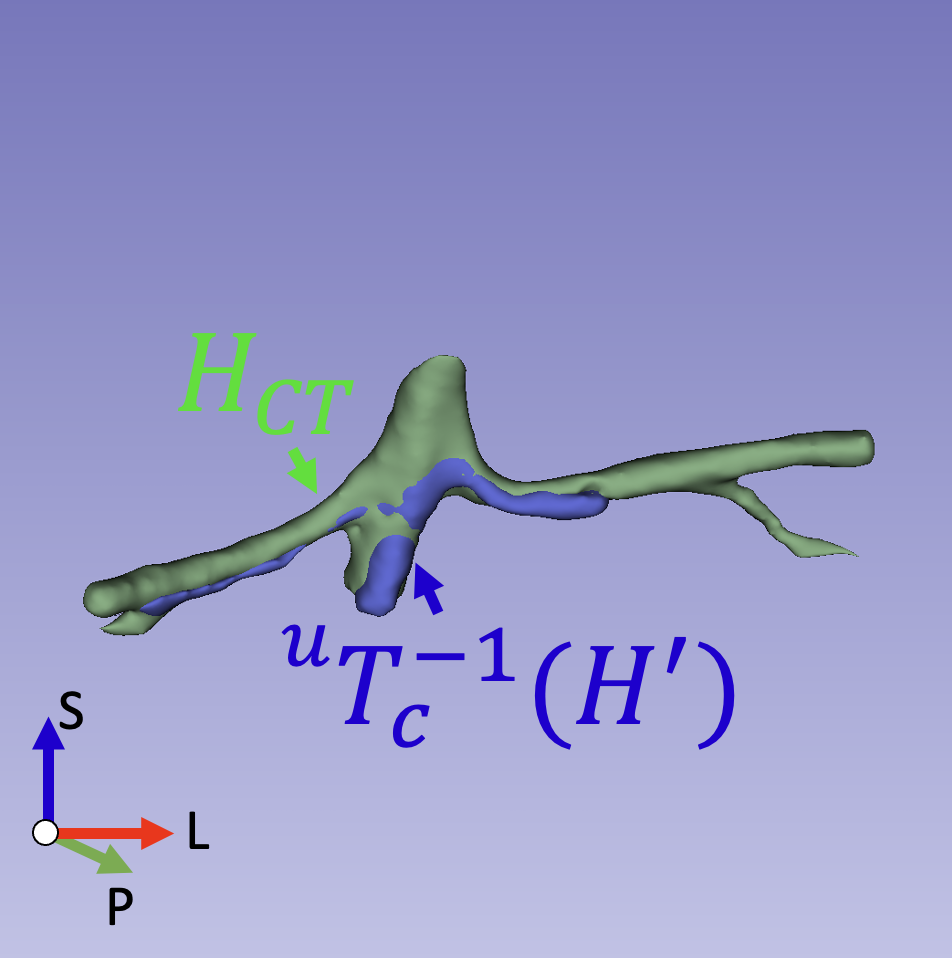}\label{subfig:post-registration}
    }
    \caption{Relative poses between the acquired HV image (blue) and the HV ground truth (green) at different stages of the coordinate mapping computation. (a) Initial images before translation (Algorithm \ref{alg:coordinate-map}, line 2). (b) Images after translating the acquired HV image $\mathcal{H}_{US}$ to $\mathcal{H}'$ (Algorithm \ref{alg:coordinate-map}, line 6). (c) Images after registration.}
    \label{fig:coordinate-mapping}
\end{figure}

The acquired 3D HV image $\mathcal{H}_{US}$ is registered with the annotation $\mathcal{H}_{CT}$ in CT to compute the crucial correspondence $^pT_c$ between the CT coordinates and physical coordinates. This coordinate mapping computation consists of three major steps: initialization, translation, and registration, as described in Algorithm \ref{alg:coordinate-map} and illustrated in Fig.~\ref{fig:coordinate-mapping}. 

Given the acquired $\mathcal{H}_{US}$ from Algorithm \ref{alg:hv_acquisition} (Fig. \ref{subfig:hv-acquisition-result}) and the ground truth HV image $\mathcal{H}_{CT}$ from CT (Fig. \ref{subfig:hv-CT-groundtruth}), the first step is to crop and resample their image arrays, so that they have the same shape and spatial resolution (line 2). The image arrays of $\mathcal{H}_{US},\mathcal{H}_{CT}$ are both 3D binary arrays, where non-zero voxels indicate the position of HV. Note that during this initialization step, the coordinate systems associated with $\mathcal{H}_{US}$ and $\mathcal{H}_{CT}$ are unchanged.

The second step is to translate the image $\mathcal{H}_{US}$ so that the centroid of the translated image $\mathcal{H}'$ coincides with the centroid of $\mathcal{H}_{CT}$ (line 4-6). This translation brings the orginally far-apart HV structures in the two images closer together, as illustrated in Figs.~\ref{subfig:pre-shift}--\ref{subfig:pre-registration}, which creates a good initial condition for the computation in the third step. 

The third step starts by computing the registration $^u T_{c}$ between the translated image $\mathcal{H'}$ and $\mathcal{H}_{CT}$ (line 8) such that the voxel values in these two images match one another, as illustrated in Fig.~\ref{subfig:post-registration}. The method for computing $^u T_c$ is the classic mutual information based registration method\cite{viola1997alignment}, which is readily available from off-the-shelf computing packages such as SimpleITK\cite{SimpleITK}. Finally, $^u T_c$ is composited with 
\begin{itemize}
    \item $\textbf{Translation}(\mathbf{g}_{US}-\mathbf{g}_{CT})$, the inverse translation of the second step, and
    \item $^p T_u$, the transformation between ultrasound frame and robot base frame (easily computable)
\end{itemize}  to become $^p T_c$--the desired transformation between CT frame and robot base frame(lines 9-10).

\begin{algorithm}[t]
    \caption{CT to Physical Coordinate Mapping}\label{alg:coordinate-map}
    \Required{$\mathcal{H}_{US}=(H,\mathbf{s},\mathbf{O},[\mathbf{a_0}, \mathbf{a_1}, \mathbf{a_2}])$, the 3D image of HV acquired from ultrasound; $\mathcal{H}_{CT}=(H_{CT},\mathbf{s}_{CT},\mathbf{O}_{CT},[\mathbf{z_0}, \mathbf{z_1}, \mathbf{z_2}])$, ground truth HV image from CT; resample target spacing $(m_0,m_1,m_2)$, target image shape $(l_0,l_1,l_2)$; $MI\_OPT(h_1,h_2)$, the registration solver; $(\mathbf{\hat{O}},[\mathbf{\hat{x}},\mathbf{\hat{y}},\mathbf{\hat{z}}])$, the origin and axes vectors for the physical coordinate system(robot base frame).}  
    
    \KwOut{A coordinate transformation map $^p T_c$ that takes a coordinate in the CT frame to the physical space(robot base frame).}
    \LinesNumbered

    \algcomments{Initialization}
    
    Resample and crop $\mathcal{H}_{US},\mathcal{H}_{CT}$ so that both have with the same spacing $(m_0,m_1,m_2)$ and image shape $(l_0,l_1,l_2)$.

    \algcomments{Translation}
    
    $\mathbf{g}_{US} \gets$ Coordinates of HV's centroid in $\mathcal{H}_{US}$.

    $\mathbf{g}_{CT} \gets$ Coordinates of HV's centroid of in $\mathcal{H}_{CT}$.

    $\mathcal{H}'\gets$ Translate $\mathcal{H}_{US}$ by $\mathbf{g}_{CT}-\mathbf{g}_{US}$.

    \algcomments{Registration}
    
    $^{u} T_{c} \gets MI\_OPT(\mathcal{H}_{CT},\mathcal{H}')$ \algcomments{$^u T_c$ maps from $\mathcal{H}_{CT}$ to $\mathcal{H}'$. } 
    
    $^p T_u\gets$ Coordinate transformation from the ultrasound image frame $(\mathbf{O},[\mathbf{a_0}, \mathbf{a_1}, \mathbf{a_2}])$ to the robot base frame $(\mathbf{\hat{O}},[\mathbf{\hat{x}},\mathbf{\hat{y}},\mathbf{\hat{z}}])$.
    
    $^p T_c\gets {^p T_u} \circ \textbf{Translation}(\mathbf{g}_{US}-\mathbf{g}_{CT}) \circ {^uT_c}$
    
    return $^p T_c$
\end{algorithm}

\subsection{Target Localization and Imaging}\label{subsec:target-localiation-imaging}

The final step of the pipeline is to have the robot move the probe to aim at the target and capture ultrasound images thereof, a.k.a., \textbf{\textit{target localization and imaging}}. Given the target coordinates $\mathbf{g}$ in CT and the coordinate mapping $^p T_c$ obtained from the previous step, we have a rough estimate $\hat{\mathbf{g}}=$$^p T_c(\mathbf{g})$ about the target's physical coordinates. For reliable target imaging, we need to refine the estimate $\hat{\mathbf{g}}$ to become a more accurate estimate $\hat{\mathbf{r}}$ through a local slice-matching procedure that utilizes the full-vessel model, the details of which are deferred to Appendix \ref{append:slice-matching-procedure}. The robot then moves the probe to step through a sequence of $N$ waypoints $\mathbf{v}_0,\mathbf{v}_1,...,\mathbf{v}_{N-1}$ on the body surface along the physical x-direction,  to sample US images in the neighborhood of $\hat{\mathbf{r}}=[\hat{r}_x,\hat{r}_y,\hat{r}_z]$. Specifically, $\mathbf{v}_i$ is defined as
$$
    \mathbf{v}_i=[\hat{r}_x-\epsilon+\frac{2i\epsilon}{N},\hat{r}_y,Z].
$$
The $\epsilon$ parameter above represents the scanning range. $Z$ is the height of body surface above the robot base frame, which is generally different from $\hat{r}_z$. We expect the target to be captured in at least one of the US images at $\mathbf{v}_0,\mathbf{v}_1,...,\mathbf{v}_{N-1}$, given that $\hat{\mathbf{r}}$ is accurate enough. 
% The entire ultrasound imaging pipeline is hence complete.

\section{Experiments}
% \tianpeng{
% Figures to add:

% \begin{itemize}
%     \item Target Imaging: Multiple figures for multi-trial results.
% \end{itemize}
% }

This section demonstrates the results of a series of experiments that evaluate the various components of our ultrasound imaging pipeline, including vessel segmentation, coordinate mapping, and most importantly target imaging.  
\footnote{
A video demo showcasing the system's operation can be found at \url{https://youtu.be/DfLY6-RNPdk}}.

The robot control source code for the following experiments can be found at \cite{Zhang_Robotic_Ultrasound_2023}. We develop the package using Python. We use the software package \texttt{ur\_rtde} by Universal Robots to issue motion commands to the UR5e manipulator, which can achieve sub-millimiter precision for end-effector positional control. The Python API of \textit{Intel\textsuperscript{\textregistered} RealSense\texttrademark~SDK 2.0} handles the sensory output from the RGB-D camera and provides the functionality to convert the RGB-D data into various formats. The UxPlay server(\url{https://github.com/FDH2/UxPlay}) manages the connection between PC and smartphone. 
The Python library of SimpleITK \cite{SimpleITK} pre-processes of CT and US images and carries out the 3D vessel registration. The AI models were trained using PyTorch framework, using the computational power of an NVIDIA A100 GPU. 

\subsection{Vessel Segmentation and Registration}\label{result:segmentation-registration}

\begin{figure}[htp]
    \centering
    \subfigure[Full-vessel]{
        \includegraphics[width = 0.8\linewidth]{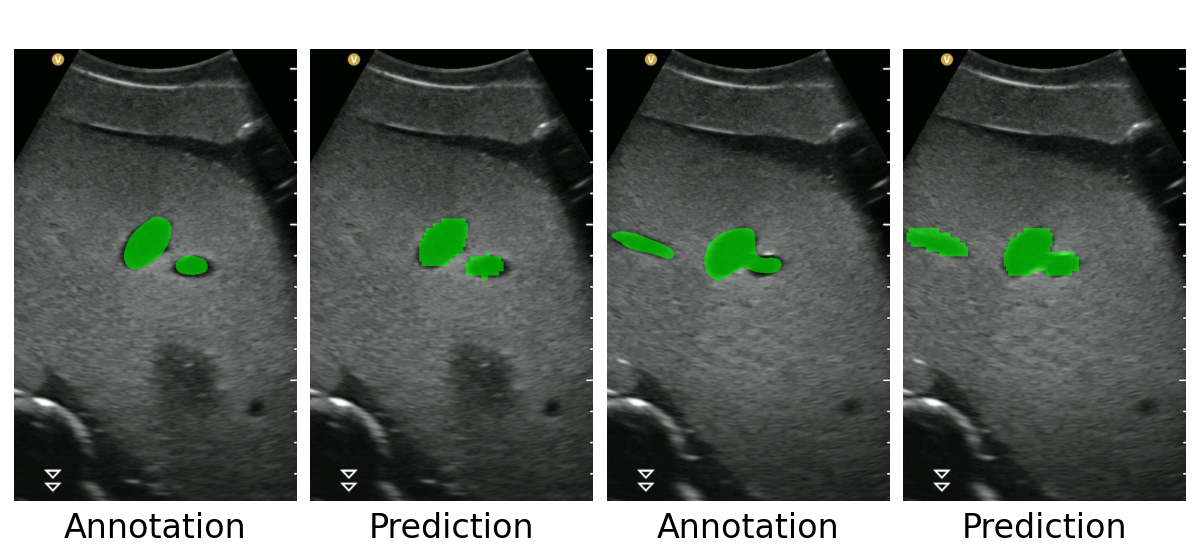}
    }
    \subfigure[Branching Point]{
        \includegraphics[width = 0.8\linewidth]{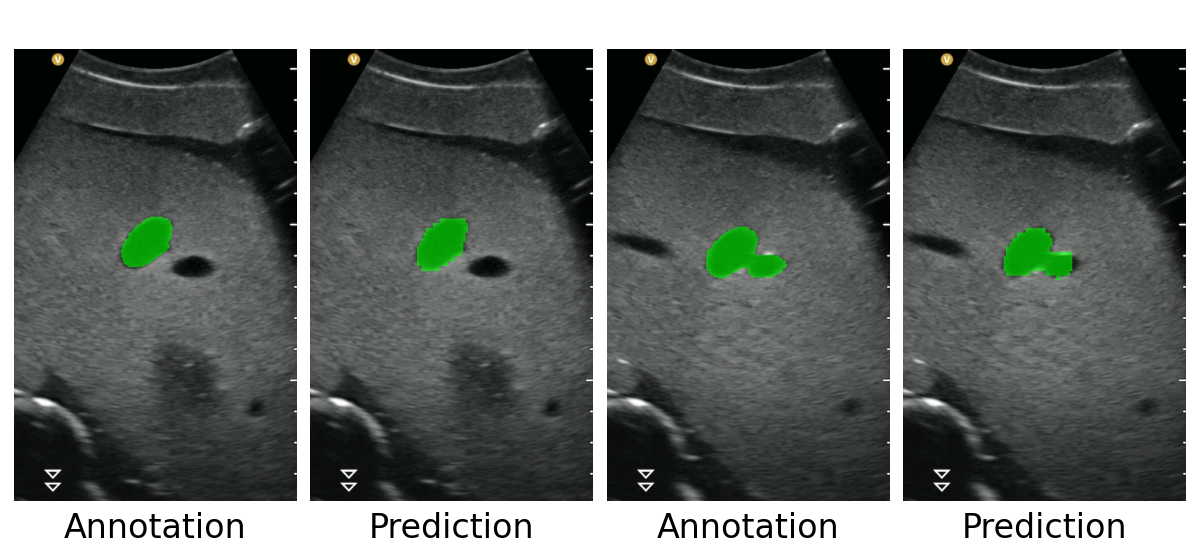}
    }
    \caption{Examples of segmentation results of the two models we trained. Images titled \textbf{Annotation} are the ground truth. Images titled \textbf{Prediction} are predicted segmentation masks returned by the two models.}\label{fig:vis_segl}
\end{figure}

\begin{figure}[htp]
    \centering
    \subfigure[Full-vessel]{\includegraphics[width = 0.75\linewidth]{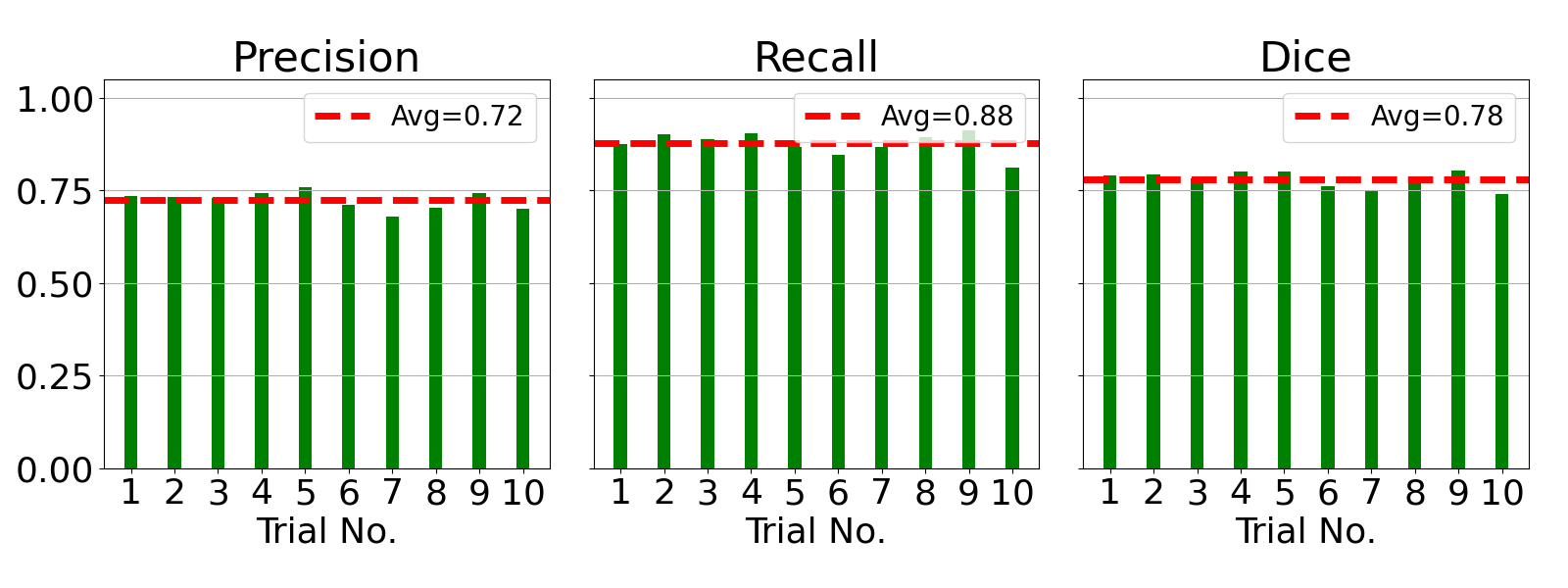}}
    \subfigure[Branching Point]{\includegraphics[width = 0.75\linewidth]{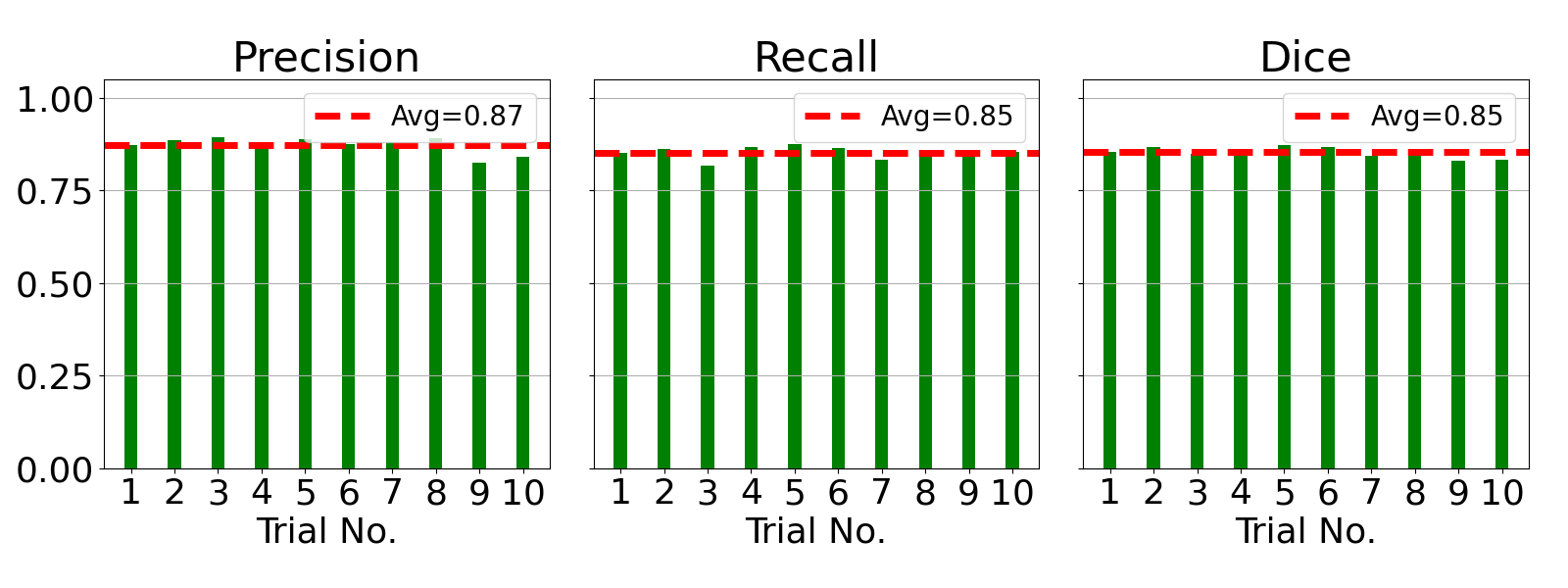}}
    \caption{Vessel segmentation performance over 10 trials. Metrics are defined in Appendix \ref{append:similarity-metrics}.}\label{fig:segmentation-bar-chart}
\end{figure}

\begin{figure}[htp]
    \centering
        \subfigure[Sample Registration Results]{\includegraphics[width = 0.90\linewidth]{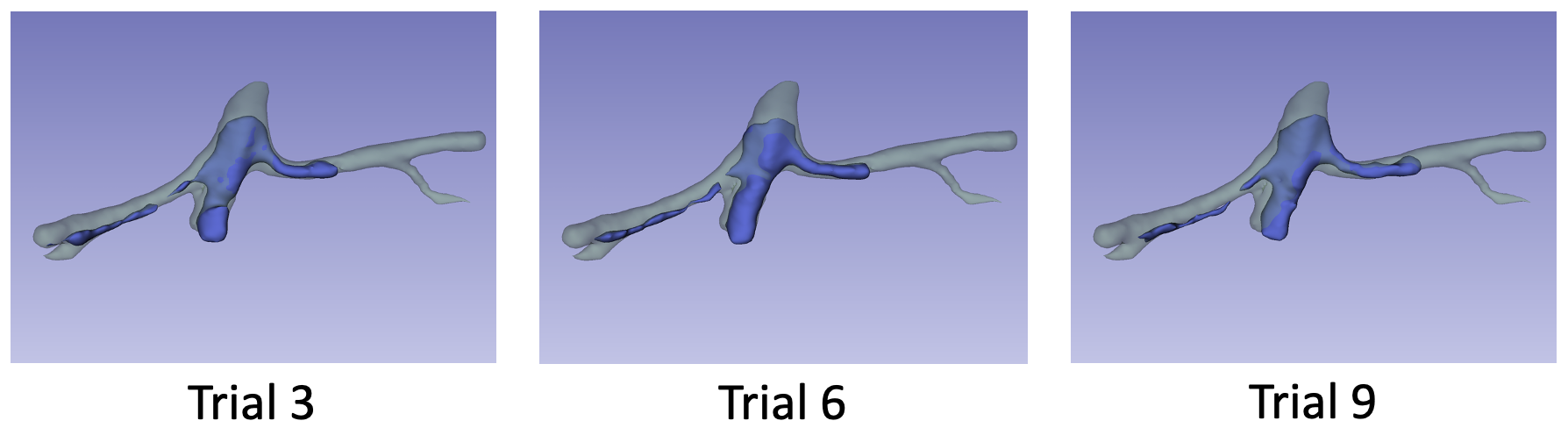}\label{subfig:registration-demo}}
        \subfigure[Image Similarity Scores of the 10 Trials]{
        \includegraphics[width = 0.90\linewidth]{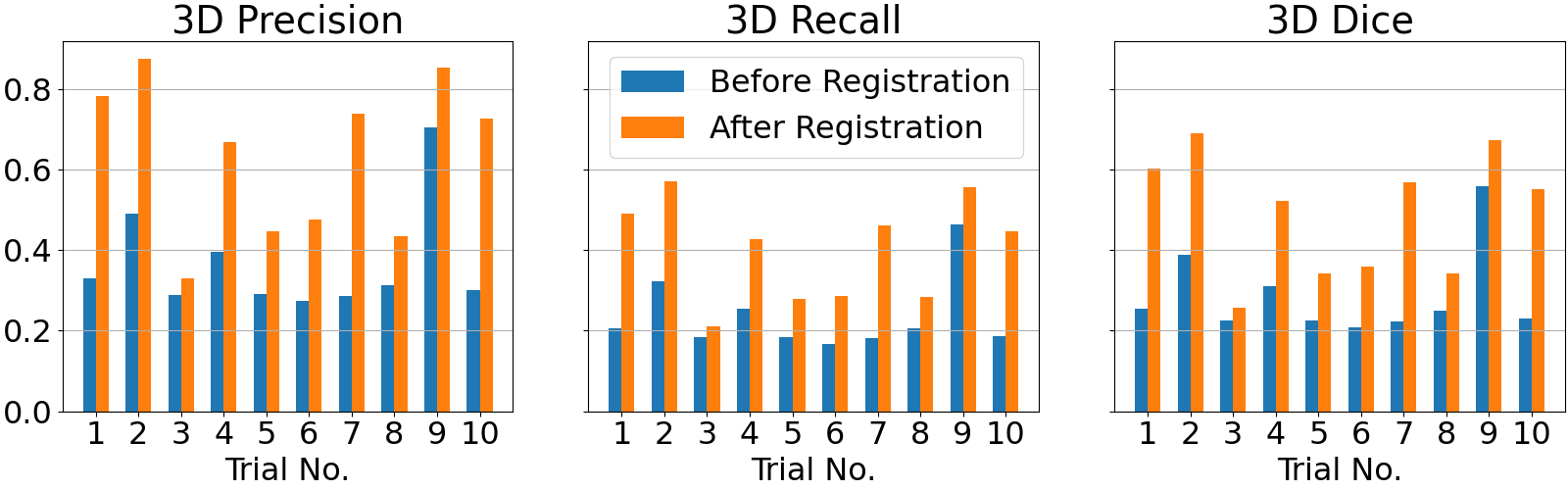}\label{fig:registration-metric}
        }
    \caption{Coordinate mapping results. (a): Sample registration results. Ground truth $\mathcal{H}_{CT}$ is labeled by transparent green pixels. Acquired HV model after registration, i.e., $^u T_c^{-1}(\mathcal{H}')$, is labeled by solid blue pixels. (b): Image similarity scores before and after registration. Blue bars: similarity between $\mathcal{H}_{CT}$ and $\mathcal{H}'$. Orange bars: similarity between $\mathcal{H}_{CT}$ and $^u T_c^{-1}(\mathcal{H}')$. Metrics are defined in Appendix \ref{append:similarity-metrics}.}\label{fig:registration-results}
\end{figure}

\begin{table}[t]
\caption{Coordinate Mapping Statistics over the 10 Trials in Fig.\ref{fig:registration-metric}}\label{table:coordinate-mapping}
\centering
\begin{tabular}{|l|c|c|}
\hline
 & Before Reg. & After Reg.\\ 
 \hline
 Ground Truth& \multicolumn{2}{c|}{$\mathcal{H}_{CT}$}\\ 
\hline
Acquired HV Image & $\mathcal{H'}$  & $^u T_c^{-1}(\mathcal{H'})$ \\ 
\hline
3D Precision(Avg.) & 0.37& 0.63 \color{green}{$\uparrow 0.26$} \\ 
\hline
3D Recall(Avg.)& 0.24& 0.40 \color{green}{$\uparrow 0.16$} \\ 
\hline
3D Dice(Avg.)& 0.29&  0.49 \color{green}{$\uparrow 0.20$}\\ 
\hline
\end{tabular}
\end{table}

This section evaluates our vessel segmentation models and the quality of US-CT registration. We run the pipeline from initial contact to coordinate mapping for 10 trials, each trial with the phantom locating at a different location on the working surface. The HV acquisition component collects 64 US images in each trial, making a testing dataset of 640 US images. We then manually annotate the HV pixels in this test set as the ground-truth labels. Fig.~\ref{fig:vis_segl} visualizes a few selected examples of side-by-side comparisons between the ground-truth annotation and model predictions. Fig.~\ref{fig:segmentation-bar-chart} reports the comprehensive segmentation performance over 10 trials in terms of precision, recall, and Dice score (as defined in Appendix \ref{append:similarity-metrics}). With an average score above 0.85 across all performance metrics, the branching point model proves to be very reliable for HV search and centralization. The full-vessel model does not score as high in precision and Dice score as the branching point model, potentially because it covers more HV pixels. 
Nevertheless, the full-vessel model does create a good enough 3D HV model for US-CT registration, as discussed below. 

Fig.~\ref{subfig:registration-demo} illustrates a few selected registrations results in the 10 trials. Results of the remaining trials are similar to those in Fig.~\ref{subfig:registration-demo}. Overall, the acquired HV models' positions match well with the ground truth after registration. Particularly, note that the positions of the three branches of hepatic veins are well-aligned.  Fig.~\ref{fig:registration-metric} and Table \ref{table:coordinate-mapping} quantitatively shows that after registration,
% , by applying the inverse of obtained US-CT registration $^u T_c^{-1}$ on the shifted HV model $H'$, 
the similarity scores between acquired HV image and CT ground truth increase noticeably.

\begin{figure*}[t]
    \centering
        \subfigure[Target]{\includegraphics[width = 0.18\linewidth]{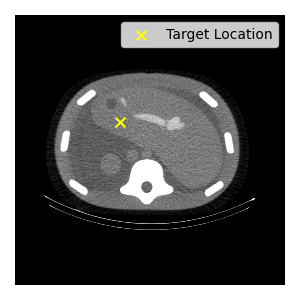}}
        \subfigure[Ultrasound Images]{\includegraphics[width = 0.7\linewidth]{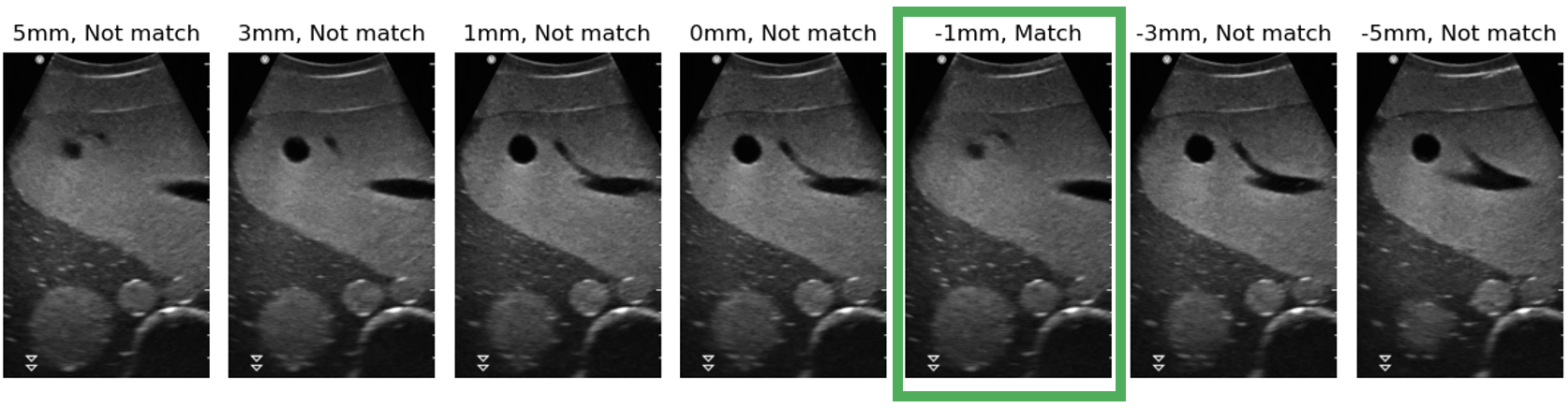}}
    \caption{An example of a target and the ultrasound images taken about the target using our method. (a) The target location  $\mathbf{g}$ in CT. (b) The sequence of US images taken about the target. The image matching the target in CT is highlighted in a green box.  Titles of the US images: the distance between the location at which the image is taken and the estimated target location $\mathbf{\hat{r}}={^p T_c}(\mathbf{g})$, also whether the US image is deemed matching the target through our manual judgement. Since there is a matching image at distance $-1$mm, the target is deemed successfully found under scanning range $\epsilon=1$mm. }\label{fig:matching-results}
\end{figure*}

% \begin{figure}[ht]
%     \centering
%         \includegraphics[width = 0.99\linewidth]{figures/coordinate_mapping/registration_metric.png}
%     \caption{3D Image Similarity Metrics Before and After Mapping in 10 ultrasound scanning experiments. }
% \end{figure}

% \tianpeng{
% Figures to add:

% \begin{itemize}
%     % \item Vessel Segmentation: Create ground truth annotation for the branching point model and plot results accordingly\textbf{[Done]}.
%     % \item Coordinate Mapping: Add 3D screenshots of pre- and post-registration results(like Figure \ref{subfig:post-registration}), for each of the 10 trials\textbf{[Done]}.
%     % \item Target imaging: Add $\pm 10$mm marker photo overlaid with the phantom body to demonstrate the scanning range[Done]
%     % \item Target imaging: Add $\pm 2$mm figure illustrating what counts as match and not match in ultrasound images[Done]
%     \item Target Imaging: Multiple figures for multi-trial results.
%     \item Target Imaging: Comparison with baseline registration method.
% \end{itemize}
% }

\subsection{Target Imaging Performance}\label{result:target-imaging}

% \tianpeng{Should we compare with a few other methods? 1) \cite{fuerst2014automatic}, multi-modality registration metric $LC^2$ to match US and CT image directly. 2) \cite{hennersperger2016towards,graumann2016robotic}, registration based on the phantom's surface, between RGB-D image and CT. }

% \tianpeng{[We need more figures in this section about target imaging performance]}

\begin{figure}[ht]
    \centering
        \includegraphics[width = 0.45\linewidth]{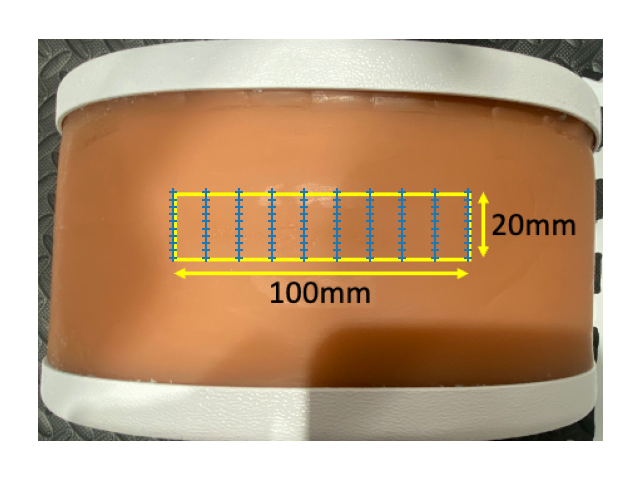}
    \caption{The distribution of target locations (small blue `+' shapes) projected on the surface of the body.}\label{fig:target-locations}
\end{figure}

This subsection demonstrates the target imaging performance of the entire pipeline. In the experiments for this subsection, we specify 100 locations in the CT image to be targeted by the ultrasound probe. These locations have the same z-coordinate, while their x-y coordinates are distributed on a 100mm by 20mm grid centered around the liver region, as illustrated in Fig.~\ref{fig:target-locations}.

The experiment to evaluate the success rate of target localization and imaging is as follows: after running the pipeline from start to the completion of coordinate mapping, for each target location we run the target localization and imaging component (Section \ref{subsec:target-localiation-imaging}) to search for and take images about the target. We determine whether the target is successfully found by manually observing whether anatomical features near the target location in CT matches one of the ultrasound images taken by the robot, and record the observations as 0-1 data points, see Fig.~ \ref{fig:matching-results}. Finally, we compute and report the success rate of finding the target among the 100 target locations under various scanning ranges $\epsilon$, the parameter introduced in Section \ref{subsec:target-localiation-imaging}. 

\begin{figure}[ht]
    \centering
    \includegraphics[width = 0.8\linewidth]{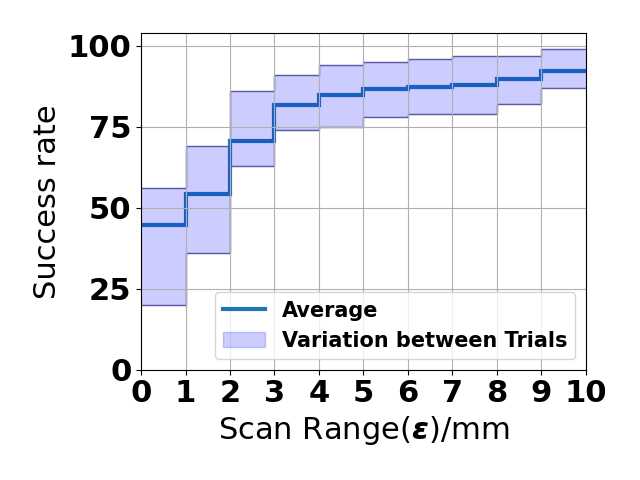}
    \caption{Success rate as a function of scan range($\epsilon$). Solid blue line: the average success rate over the trials. Light-blue shaded area: variation in success rate over the trials. The upper/lower boundary of the shaded area corresponds to the highest/lowest success rate at each $\epsilon$ value across the trials.  }
    \label{fig:success_rate}
\end{figure}

We repeat the experiment above for 5 trials, each trial with a different initial position of the phantom on the working surface. Fig.~\ref{fig:success_rate} visualizes the success rate of finding the target as a function of scanning range $\epsilon$, showing the average success rate and variation over different trials. Table \ref{table:success_rate} displays a few key entries of the data in Fig.~\ref{fig:success_rate}. As the scanning range increases, the target is more likely to be included in the set of images taken, and therefore the success rate increases as expected. The results show that, on average, over 44\% of the target locations are within 1mm of the estimated location $\mathbf{\hat{r}}$ (introduced in Section \ref{subsec:target-localiation-imaging}), and near 90\% of all targets are found within the 9mm scanning range. 

\begin{table}[htp]
\caption{Success Rates at Different Scan Ranges($\epsilon$) Over 5 Trials}\label{table:success_rate}
\centering
\begin{tabular}{|l|l|l|l|l|}
\hline
Scan Range($\epsilon$) & $\pm$1mm & $\pm$3mm & $\pm$5mm & $\pm$9mm \\ \hline
Avg. Success Rate & 44.8\%  & 70.6\%  & 84.8\%  & 89.8\%  \\ \hline
Highest Success Rate & 56.0\%  & 86.0\%  & 94.0\%  & 97.0\%  \\ 
\hline 
Lowest Success Rate & 20.0\%  & 63.0\%  & 75.0\%  & 82.0\%  \\ 
\hline
\end{tabular}
\end{table}

% \begin{definition}
%     Let $\mathbf{\hat{p}} = [\hat{p}_x, \hat{p}_y, \hat{p}_z]$, $\mathbf{p}=[p_x,p_y,p_z]$ be the physical coordinates of the probe and the target respectively. The target is deemed successfully located by the probe when the following conditions are both satisfied.
%     \begin{itemize}
%         \item[a)] $|\hat{p}_x-p_x|<\epsilon$, $\epsilon$ is a pre-specified small constant.
%         \item[b)] $|\hat{p}_y-p_y|<w/2$, $w$ is the physical width of the ultrasound field of view.
%     \end{itemize}
% \end{definition}

\section{Discussion}
Our pilot study is still limited and has yet to consider factors including patients' respiratory and cardiac movements, optimal transducer pressure, titling angles, as well as variations in patient positioning, such as lying on their side or flat. These factors will be comprehensively addressed in subsequent studies.

While our proposed method focused on liver follow-up scans, our framework can be easily extended to other organs. We are developing various deep-learning modules based on specific target regions and tasks. Particularly, for tasks involving ultrasound images, we plan to employ a foundation model combined with task-specific adapters. Our foundation model, built upon MGH ultrasound images, is currently under development. Once finalized, each specialized task is expected to yield improved performance, even with limited datasets.

Another aspect of our research pertains to enhancing communication between the robot and the patient, which directly influences the assurance of data quality and is crucial for precise diagnosis. Assessing quality assurance will involve the analysis of camera and ultrasound images. If the quality is unsatisfactory, the robot can autonomously adjust its position or receive guidance from the AI model to optimize patient positioning. Communication will leverage local large language models (LLMs) such as Llama3-8B, integrated with a Langchain for text-to-speech capabilities. Additionally, we plan to integrate large multi-modal models (LMMs), incorporating both language and vision models, in the near future.

\section{Conclusion}
The paper proposed an autonomous robotic ultrasound system designed specifically for liver follow-up scans within local communities. The system incorporates three primary functionalities: (i) initial robot contact to the surface using RGB-D camera, (ii) precise coordinate mapping between the CT image and the robot base frame using 3D US-CT registration of hepatic veins, and (iii) targeted ultrasound scanning guided by coordinate mapping and local slice matching. Pilot experiments using an abdomen phantom demonstrated the high-quality registration and successful ultrasound scans of automatically localized targets. Future research will involve expanding the study to human subjects including communication with human and robot by large multi-modal models. 

% \section*{Acknowledgments}
% This should be a simple paragraph 

\bibliographystyle{IEEEtran}
\bibliography{references}

\begin{thebibliography}{10}
\providecommand{\url}[1]{#1}
\csname url@samestyle\endcsname
\providecommand{\newblock}{\relax}
\providecommand{\bibinfo}[2]{#2}
\providecommand{\BIBentrySTDinterwordspacing}{\spaceskip=0pt\relax}
\providecommand{\BIBentryALTinterwordstretchfactor}{4}
\providecommand{\BIBentryALTinterwordspacing}{\spaceskip=\fontdimen2\font plus
\BIBentryALTinterwordstretchfactor\fontdimen3\font minus
  \fontdimen4\font\relax}
\providecommand{\BIBforeignlanguage}[2]{{%
\expandafter\ifx\csname l@#1\endcsname\relax
\typeout{** WARNING: IEEEtran.bst: No hyphenation pattern has been}%
\typeout{** loaded for the language `#1'. Using the pattern for}%
\typeout{** the default language instead.}%
\else
\language=\csname l@#1\endcsname
\fi
#2}}
\providecommand{\BIBdecl}{\relax}
\BIBdecl

\bibitem{saraogi2015lung}
A.~Saraogi, ``{Lung ultrasound: Present and future},'' \emph{Lung India:
  Official Organ of Indian Chest Society}, vol.~32, no.~3, p. 250, 2015.

\bibitem{berg2006operator}
W.~A. Berg, J.~D. Blume, J.~B. Cormack, and E.~B. Mendelson, ``Operator
  dependence of physician-performed whole-breast us: lesion detection and
  characterization,'' \emph{Radiology}, vol. 241, no.~2, pp. 355--365, 2006.

\bibitem{jiang2020automatic}
Z.~Jiang, M.~Grimm, M.~Zhou, Y.~Hu, J.~Esteban, and N.~Navab, ``Automatic
  force-based probe positioning for precise robotic ultrasound acquisition,''
  \emph{IEEE Transactions on Industrial Electronics}, vol.~68, no.~11, pp.
  11\,200--11\,211, 2020.

\bibitem{sonko2022machine}
M.~L. Sonko, T.~C. Arnold, and I.~A. Kuznetsov, ``{Machine Learning in Point of
  Care Ultrasound},'' \emph{POCUS Journal}, vol.~7, no. Kidney, pp. 78--87,
  2022.

\bibitem{kim2023point}
K.~Kim, F.~Macruz, D.~Wu, C.~Bridge, S.~McKinney, A.~A. Al~Saud, E.~Sharaf,
  A.~Pely, P.~Danset, T.~Duffy \emph{et~al.}, ``{Point-of-care AI-assisted
  stepwise ultrasound pneumothorax diagnosis},'' \emph{Physics in Medicine and
  Biology}, vol.~68, no.~20, p. 205013, 2023.

\bibitem{von2021medical}
F.~von Haxthausen, S.~B{\"o}ttger, D.~Wulff, J.~Hagenah,
  V.~Garc{\'\i}a-V{\'a}zquez, and S.~Ipsen, ``Medical robotics for ultrasound
  imaging: current systems and future trends,'' \emph{Current robotics
  reports}, vol.~2, pp. 55--71, 2021.

\bibitem{li2023overview}
K.~Li, Y.~Xu, and M.~Q.-H. Meng, ``An overview of systems and techniques for
  autonomous robotic ultrasound acquisitions,'' \emph{IEEE Transactions on
  Medical Robotics and Bionics}, vol.~3, no.~2, pp. 510--524, 2021.

\bibitem{jiang2023robotic}
Z.~Jiang, S.~E. Salcudean, and N.~Navab, ``Robotic ultrasound imaging:
  State-of-the-art and future perspectives,'' \emph{Medical image analysis}, p.
  102878, 2023.

\bibitem{graumann2016robotic}
C.~Graumann, B.~Fuerst, C.~Hennersperger, F.~Bork, and N.~Navab, ``Robotic
  ultrasound trajectory planning for volume of interest coverage,'' in
  \emph{2016 IEEE international conference on robotics and automation
  (ICRA)}.\hskip 1em plus 0.5em minus 0.4em\relax IEEE, 2016, pp. 736--741.

\bibitem{hennersperger2016towards}
C.~Hennersperger, B.~Fuerst, S.~Virga, O.~Zettinig, B.~Frisch, T.~Neff, and
  N.~Navab, ``Towards mri-based autonomous robotic us acquisitions: a first
  feasibility study,'' \emph{IEEE transactions on medical imaging}, vol.~36,
  no.~2, pp. 538--548, 2016.

\bibitem{mylonas2013autonomous}
G.~P. Mylonas, P.~Giataganas, M.~Chaudery, V.~Vitiello, A.~Darzi, and G.-Z.
  Yang, ``Autonomous efast ultrasound scanning by a robotic manipulator using
  learning from demonstrations,'' in \emph{2013 IEEE/RSJ International
  Conference on Intelligent Robots and Systems}.\hskip 1em plus 0.5em minus
  0.4em\relax IEEE, 2013, pp. 3251--3256.

\bibitem{ramalhinho2022deep}
J.~Ramalhinho, B.~Koo, N.~Monta{\~n}a-Brown, S.~U. Saeed, E.~Bonmati,
  K.~Gurusamy, S.~P. Pereira, B.~Davidson, Y.~Hu, and M.~J. Clarkson, ``Deep
  hashing for global registration of untracked 2d laparoscopic ultrasound to
  ct,'' \emph{International Journal of Computer Assisted Radiology and
  Surgery}, vol.~17, no.~8, pp. 1461--1468, 2022.

\bibitem{ning2021autonomic}
G.~Ning, X.~Zhang, and H.~Liao, ``Autonomic robotic ultrasound imaging system
  based on reinforcement learning,'' \emph{IEEE Transactions on Biomedical
  Engineering}, vol.~68, no.~9, pp. 2787--2797, 2021.

\bibitem{allen2018healthcare}
A.~M. Allen, H.~K. Van~Houten, L.~R. Sangaralingham, J.~A. Talwalkar, and R.~G.
  McCoy, ``{Healthcare cost and utilization in nonalcoholic fatty liver
  disease: real-world data from a large US claims database},''
  \emph{Hepatology}, vol.~68, no.~6, pp. 2230--2238, 2018.

\bibitem{nakadate2010implementation}
R.~Nakadate, J.~Solis, A.~Takanishi, E.~Minagawa, M.~Sugawara, and K.~Niki,
  ``Implementation of an automatic scanning and detection algorithm for the
  carotid artery by an assisted-robotic measurement system,'' in \emph{2010
  IEEE/RSJ International Conference on Intelligent Robots and Systems}.\hskip
  1em plus 0.5em minus 0.4em\relax IEEE, 2010, pp. 313--318.

\bibitem{mustafa2013development}
A.~S.~B. Mustafa, T.~Ishii, Y.~Matsunaga, R.~Nakadate, H.~Ishii, K.~Ogawa,
  A.~Saito, M.~Sugawara, K.~Niki, and A.~Takanishi, ``Development of robotic
  system for autonomous liver screening using ultrasound scanning device,'' in
  \emph{2013 IEEE international conference on robotics and biomimetics
  (ROBIO)}.\hskip 1em plus 0.5em minus 0.4em\relax IEEE, 2013, pp. 804--809.

\bibitem{merouche2015robotic}
S.~Merouche, L.~Allard, E.~Montagnon, G.~Soulez, P.~Bigras, and G.~Cloutier,
  ``A robotic ultrasound scanner for automatic vessel tracking and
  three-dimensional reconstruction of b-mode images,'' \emph{IEEE transactions
  on ultrasonics, ferroelectrics, and frequency control}, vol.~63, no.~1, pp.
  35--46, 2015.

\bibitem{virga2016automatic}
S.~Virga, O.~Zettinig, M.~Esposito, K.~Pfister, B.~Frisch, T.~Neff, N.~Navab,
  and C.~Hennersperger, ``Automatic force-compliant robotic ultrasound
  screening of abdominal aortic aneurysms,'' in \emph{2016 IEEE/RSJ
  international conference on intelligent robots and systems (IROS)}.\hskip 1em
  plus 0.5em minus 0.4em\relax IEEE, 2016, pp. 508--513.

\bibitem{kojcev2017reproducibility}
R.~Kojcev, A.~Khakzar, B.~Fuerst, O.~Zettinig, C.~Fahkry, R.~DeJong,
  J.~Richmon, R.~Taylor, E.~Sinibaldi, and N.~Navab, ``On the reproducibility
  of expert-operated and robotic ultrasound acquisitions,'' \emph{International
  journal of computer assisted radiology and surgery}, vol.~12, pp. 1003--1011,
  2017.

\bibitem{chatelain20153d}
P.~Chatelain, A.~Krupa, and N.~Navab, ``3d ultrasound-guided robotic steering
  of a flexible needle via visual servoing,'' in \emph{2015 IEEE International
  Conference on Robotics and Automation (ICRA)}.\hskip 1em plus 0.5em minus
  0.4em\relax IEEE, 2015, pp. 2250--2255.

\bibitem{kojcev2016dual}
R.~Kojcev, B.~Fuerst, O.~Zettinig, J.~Fotouhi, S.~C. Lee, B.~Frisch, R.~Taylor,
  E.~Sinibaldi, and N.~Navab, ``Dual-robot ultrasound-guided needle placement:
  closing the planning-imaging-action loop,'' \emph{International journal of
  computer assisted radiology and surgery}, vol.~11, pp. 1173--1181, 2016.

\bibitem{salcudean1999robot}
S.~E. Salcudean, G.~Bell, S.~Bachmann, W.-H. Zhu, P.~Abolmaesumi, and P.~D.
  Lawrence, ``Robot-assisted diagnostic ultrasound--design and feasibility
  experiments,'' in \emph{Medical Image Computing and Computer-Assisted
  Intervention--MICCAI’99: Second International Conference, Cambridge, UK,
  September 19-22, 1999. Proceedings 2}.\hskip 1em plus 0.5em minus 0.4em\relax
  Springer, 1999, pp. 1062--1071.

\bibitem{essomba2012design}
T.~Essomba, L.~Nouaille, M.~Laribi, G.~Poisson, and S.~Zeghloul, ``Design
  process of a robotized tele-echography system,'' \emph{Applied Mechanics and
  Materials}, vol. 162, pp. 384--393, 2012.

\bibitem{von2020robotized}
F.~Von~Haxthausen, J.~Hagenah, M.~Kaschwich, M.~Kleemann,
  V.~Garc{\'\i}a-V{\'a}zquez, and F.~Ernst, ``Robotized ultrasound imaging of
  the peripheral arteries--a phantom study,'' in \emph{Current Directions in
  Biomedical Engineering}, vol.~6, no.~1.\hskip 1em plus 0.5em minus
  0.4em\relax De Gruyter, 2020, p. 20200033.

\bibitem{jiang2021autonomous}
Z.~Jiang, Z.~Li, M.~Grimm, M.~Zhou, M.~Esposito, W.~Wein, W.~Stechele,
  T.~Wendler, and N.~Navab, ``Autonomous robotic screening of tubular
  structures based only on real-time ultrasound imaging feedback,'' \emph{IEEE
  Transactions on Industrial Electronics}, vol.~69, no.~7, pp. 7064--7075,
  2021.

\bibitem{zettinig2016toward}
O.~Zettinig, B.~Fuerst, R.~Kojcev, M.~Esposito, M.~Salehi, W.~Wein,
  J.~Rackerseder, E.~Sinibaldi, B.~Frisch, and N.~Navab, ``Toward real-time 3d
  ultrasound registration-based visual servoing for interventional
  navigation,'' in \emph{2016 IEEE International Conference on Robotics and
  Automation (ICRA)}.\hskip 1em plus 0.5em minus 0.4em\relax IEEE, 2016, pp.
  945--950.

\bibitem{langsch2019robotic}
F.~Langsch, S.~Virga, J.~Esteban, R.~G{\"o}bl, and N.~Navab, ``Robotic
  ultrasound for catheter navigation in endovascular procedures,'' in
  \emph{2019 IEEE/RSJ International Conference on Intelligent Robots and
  Systems (IROS)}.\hskip 1em plus 0.5em minus 0.4em\relax IEEE, 2019, pp.
  5404--5410.

\bibitem{fuerst2014automatic}
B.~Fuerst, W.~Wein, M.~M{\"u}ller, and N.~Navab, ``Automatic ultrasound--mri
  registration for neurosurgery using the 2d and 3d lc2 metric,'' \emph{Medical
  image analysis}, vol.~18, no.~8, pp. 1312--1319, 2014.

\bibitem{haque2016automated}
H.~Haque, Y.~Omi, L.~Rusk{\'o}, P.~Annangi, and O.~Kazuyuki, ``Automated
  registration of 3d ultrasound and ct/mr images for liver,'' in \emph{2016
  IEEE International Ultrasonics Symposium (IUS)}.\hskip 1em plus 0.5em minus
  0.4em\relax IEEE, 2016, pp. 1--4.

\bibitem{he2023robust}
B.~He, S.~Zhao, Y.~Dai, J.~Wu, H.~Luo, J.~Guo, Z.~Ni, T.~Wu, F.~Kuang, H.~Jiang
  \emph{et~al.}, ``A robust and automatic ct-3d ultrasound registration method
  based on segmentation, context, and edge hybrid metric,'' \emph{Medical
  Physics}, 2023.

\bibitem{ronneberger2015u}
O.~Ronneberger, P.~Fischer, and T.~Brox, ``U-net: Convolutional networks for
  biomedical image segmentation,'' in \emph{Medical Image Computing and
  Computer-Assisted Intervention--MICCAI 2015: 18th International Conference,
  Munich, Germany, October 5-9, 2015, Proceedings, Part III 18}.\hskip 1em plus
  0.5em minus 0.4em\relax Springer, 2015, pp. 234--241.

\bibitem{keller2023skin}
M.~Keller, K.~Werling, S.~Shin, S.~Delp, S.~Pujades, C.~K. Liu, and M.~J.
  Black, ``From skin to skeleton: Towards biomechanically accurate 3d digital
  humans,'' \emph{ACM Transactions on Graphics (TOG)}, vol.~42, no.~6, pp.
  1--12, 2023.

\bibitem{finocchi2017co}
R.~Finocchi, F.~Aalamifar, T.~Y. Fang, R.~H. Taylor, and E.~M. Boctor,
  ``Co-robotic ultrasound imaging: A cooperative force control approach,'' in
  \emph{Medical Imaging 2017: Image-Guided Procedures, Robotic Interventions,
  and Modeling}, vol. 10135.\hskip 1em plus 0.5em minus 0.4em\relax SPIE, 2017,
  pp. 270--280.

\bibitem{viola1997alignment}
P.~Viola and W.~M. Wells~III, ``Alignment by maximization of mutual
  information,'' \emph{International journal of computer vision}, vol.~24,
  no.~2, pp. 137--154, 1997.

\bibitem{SimpleITK}
\BIBentryALTinterwordspacing
B.~Lowekamp, D.~Chen, L.~Ibanez, and D.~Blezek, ``The design of simpleitk,''
  \emph{Frontiers in Neuroinformatics}, vol.~7, 2013. [Online]. Available:
  \url{https://www.frontiersin.org/articles/10.3389/fninf.2013.00045}
\BIBentrySTDinterwordspacing

\bibitem{Zhang_Robotic_Ultrasound_2023}
\BIBentryALTinterwordspacing
T.~Zhang, M.~Haitong, and K.~Sekeun, ``{Robotic Ultrasound},'' Oct. 2023.
  [Online]. Available:
  \url{https://github.com/lina-robotics-lab/robotic-ultrasound}
\BIBentrySTDinterwordspacing

\end{thebibliography}
\appendices

% \section{Accessing Real-time Ultrasound Image on computers}\label{append:US2PC}
% At the time of this work, the GE VScan Air CL probe supports data streaming to smart phones and tablets, but does not support direct streaming onto computers.  Therefore, we had to first stream the probe data onto a smart phone, then relay the data to the computer via a USB connection. We discuss the technical procedure below...

% \section{Body Segmentation through Color Thresholding}\label{append:body-segmentation}
% ...

\section{Similarity Metrics for Binary Arrays}\label{append:similarity-metrics}

Let $Y$ and $\tilde{Y}$ be binary multi-dimensional arrays with the same size. Assume $\tilde{Y}$ is the ground truth, and $Y$ is a prediction about $\tilde{Y}$. Define $|Y|$ as the L1 norm of $Y$, which is the same as the number of non-zero entries in $Y$. Let $Y\cap \tilde{Y}$ be the matrix resulting from entry-wise and-operation between $Y$ and $\tilde{Y}$.

We define the following metrics that measures the similarity between the ground truth $\tilde{Y}$ and prediction $Y$:

\textbf{Precision}$= \frac{|\tilde{Y}\cap Y|}{|Y|}$. 
\textbf{Recall}$= \frac{|\tilde{Y}\cap Y|}{|\tilde{Y}|}$.
\textbf{Dice Score}$= \frac{2|\tilde{Y}\cap Y|}{|Y|+|\tilde{Y}|}$.

\section{Slice Matching Procedure}\label{append:slice-matching-procedure}

In what follows, we first explain the necessity for local slice-matching by briefly discussing the US image's sensitivity to the probe movement in physical x-y directions, then present the target localization and imaging algorithm with the slice-matching procedure. 

\textit{Notations.} Given the target location $\mathbf{g}$ in CT coordinates, let's assume $\mathbf{p}=[p_x,p_y,p_z]$ are true coordinates of $\mathbf{g}$ in the physical space. Let $\mathbf{\hat{g}}=[\hat{g}_x,\hat{g}_y,\hat{g}_z]={^p T_c}(\mathbf{g})$ be an estimate of the target's physical location $\mathbf{p}$, where $^p T_c$ is obtained from Algorithm \ref{alg:coordinate-map} in the previous step.

\subsubsection{US Image Sensitivity to x-y Directional Movements} 
\begin{figure}[htp]
    \centering
    \subfigure[Probe moves in physical x-direction by \underline{\textbf{1mm}}  increment ]{
        \includegraphics[width = 1\linewidth]{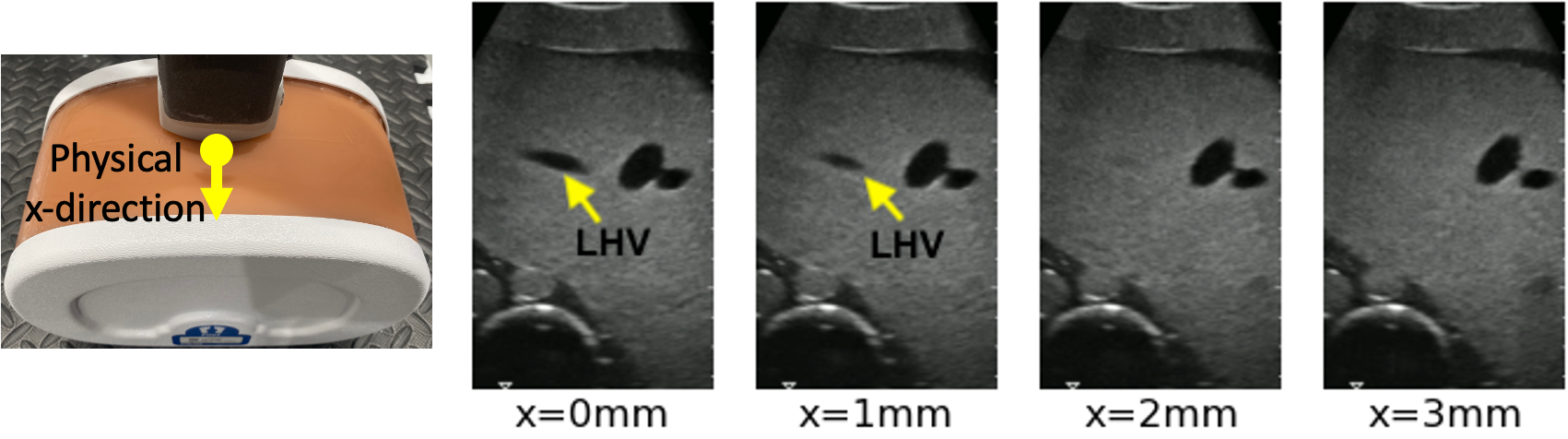}\label{subfig:x-sensitivity}
    }
    \subfigure[Probe moves in physical y-direction by \underline{\textbf{10mm}} increment]{
        \includegraphics[width = 1\linewidth]{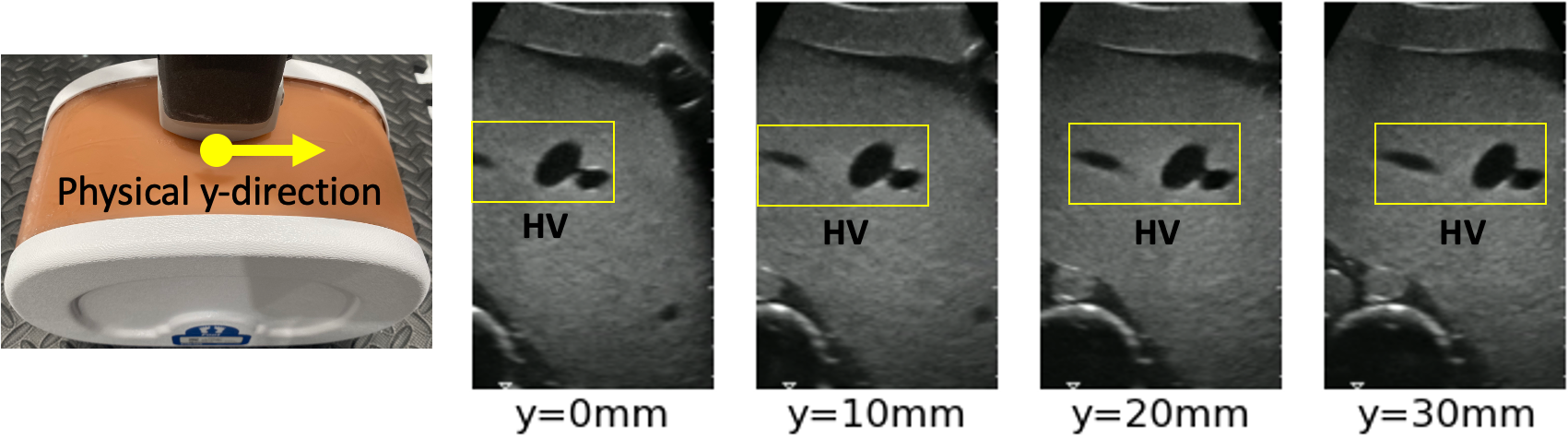}\label{subfig:y-sensitivity}
    }    
    \caption{The features in the ultrasound image are much more sensitive to probe movement in the physical x-axis than in the y-axis. In (a), the left hepatic vein (labeled \textbf{LHV}) disappears from the image after the probe moves in the x-direction for 3mm. In (b), as the probe moves in the y-direction, the hepatic vein features (labeled \textbf{HV}) remain largely invariant except for a shift in the image.
    }\label{fig:xy-sensitivity}
\end{figure}

The US images are much more sensitive to probe movements in physical x-direction than y-direction on the body's surface. In our experiment setup, the US probe orientation is fixed so that the US images are always aligned with the axial plane. When the probe moves in the x-direction, the US image switches between different axial slices, and the image features could be noticeably different even if the probe moves only for 1mm (see Fig.~\ref{subfig:x-sensitivity}). Extensive observations have indicate the x-component estimate $\hat{g}_x$ is typically not accurate enough to be within $1$mm from $p_x$. In comparison, the width of the ultrasound field of view $\approx$ 80mm, meaning the probe can deviate for up to $40$mm from the true target location in the physical y-direction and still capture the target location within the field of view (see Fig.~\ref{subfig:y-sensitivity}), and the y-component estimation error $|\hat{g}_y-p_y|$ is typically far smaller than this 40mm margin.

% To better describe and elucidate the method for this step, it is worth giving a more rigorous definition of the goal of \textbf{\textit{target localization}} as follows.  Given $\mathbf{g}$, the target location specified in the CT coordinates, let's assume $\mathbf{p}=[p_x,p_y,p_z]$ are true coordinates of $\mathbf{g}$ in the physical space. The robot must position the probe to aim at a location in the body that is as close to $\mathbf{p}$ as possible to ensure the ultrasound image contains the target. Since the probe has sufficient viewing depth to penetrate the entire liver, the target would be visible in the ultrasound image as long as two conditions are satisfied:

% \begin{itemize}
%     \item[(1)] The physical x- and y-coordinates of the probe match those of $\mathbf{p}$.
%     \item[(2)] The probe stays in contact with the body.
% \end{itemize}
% Condition (2) can be easily achieved by monitoring the contact force and restricting the range of probe location so the probe does not move away from the body. Therefore, the main goal is to achieve condition (1). 

\subsubsection{Target Imaging Algorithm with Slice-matching} The discussions above establishes that directly moving the probe to aim at $\mathbf{\hat{g}}$ does not guarantee capturing the target in the ultrasound image. Although the y-coordinate estimate $\hat{g}_y$ can be a useful reference, the physical x-coordinates of the target need to be more precisely determined. With this in mind, we develop Algorithm \ref{alg:localization-imaging} for target localization and imaging.
% Figure \ref{fig:target_localization} illustrates the locations of some key entities in Algorithm \ref{alg:localization-imaging} on the body's surface, including $\mathbf{p}_{branch},~ \mathbf{\hat{g}},~\mathbf{g'},~\mathbf{\hat{r}},$ and the slice matching waypoints. 
The location $\mathbf{p}_{branch}=[p_{cx},p_{cy},Z]$ reached in HV localization serves as the starting location of the probe. The z-coordinate of the probe is fixed hereafter and the probe only moves on the frontal plane(physical x-y plane). The robot moves the probe to $\mathbf{g}'=[\hat{g}_x,p_{cy},Z]$, which is supposedly close to the axial plane containing the target (line 2). The robot then performs a local slice matching procedure along the x-axis centered at $\hat{g}_x$, comparing the HV observed in the ultrasound images with the HV in the axial CT slice containing the target (line 4-16). The x-coordinate of the waypoint with the highest matching score ($Q_x$) becomes x-coordinate the matched axial slice and $\hat{\mathbf{r}}=[Q_x, \hat{g}_y,Z]$ becomes the refined estimate of the target's physical coordinates. Finally, the robot moves the probe to location $\hat{\mathbf{r}}$ (line 17), then take a sequence of images about the target along the physical x-axis (line 18), completing the entire ultrasound imaging pipeline.

\begin{algorithm}[htp]
    \caption{Target Localization and Imaging with Slice-matching}\label{alg:localization-imaging}
    \Required{Model $\mathcal{M}_{F}$; HV ground truth $\mathcal{H}_{CT}$; target coordinates $\mathbf{g}$; $^p T_c$; $\mathbf{p}_{branch}=[p_{cx},p_{cy},Z]$; local search distance $S$, number of waypoints $n$; scanning range $\epsilon$.}  
    
    \KwOut{Ultrasound images about the target.}
    \LinesNumbered

    Estimated target location in physical space $\mathbf{\hat{g}}=[\hat{g}_x,\hat{g}_y,\hat{g}_z]\gets {^p T_{c}}(\mathbf{g})$
        
    Move probe to $\mathbf{g}'=[\hat{g}_x,p_{cy},Z]$.

    % \algcomments{Slice-matching Starts}
        
    Generate waypoints $\mathcal{S}$: $\mathcal{S} \gets \{\mathbf{s}_i=[x_i,p_{cy},Z]| x_i\in [\hat{g}_x-S/2,\hat{g}_x+S/2],i=0,1,2,...,n-1, |x_i-x_{i-1}|=S/(n-1)\}$

    $\mathcal{S}\gets \mathcal{S}\cup \{[\hat{g}_x,p_{cy},Z]\}$

    $top\gets -\inf$, $Q_x\gets \hat{g}_x$

    $\tilde{Y}\gets$ The axial slice in $\mathcal{H}_{CT}$ containing target $\mathbf{g}$.

    \For{$\mathbf{s}=[s_x,s_y,s_z]\in \mathcal{S}$}{
        Move probe to $\mathbf{s}$. 
        Take ultrasound image $img$. Compute HV segmentation mask $Y \gets \mathcal{M}_F(img)$.

        $sc\gets score(\tilde{Y},Y)$

        \If{$sc>top$}{
            $top\gets sc$, $Q_x\gets s_x$
        }
    }

    % \algcomments{Slice-matching Ends}
    
    $\hat{\mathbf{r}}:=[\hat{r}_x,\hat{r}_y,\hat{r}_z]\gets[Q_x, \hat{g}_y,\hat{g}_z]$.

    Take a sequence of $N$ ultrasound images in the neighborhood of $\hat{\mathbf{r}}$ along the physical x-axis with scanning range $\epsilon$.
    % Waypoints to sample = $\{\mathbf{v}_i=[\hat{r}_x-\epsilon+\frac{2i\epsilon}{N}, \hat{r}_y,Z]|i=0,1,...,N-1\}$
\end{algorithm}

% \begin{figure}[htp]
%     \centering
%     \subfigure[Alg.\ref{alg:localization-imaging} lines 1-2: Relationship between $\mathbf{p}_{branch}$, $\hat{\mathbf{g}}$, and $\mathbf{g'}$]{
%         \includegraphics[width = 0.46\linewidth]{figures/target_localization/localization_0.png}\label{subfig:localization_0}
%     }
%     \subfigure[Alg. \ref{alg:localization-imaging} line 4: Slice matching waypoints are equally spaced along x-axis centered $\mathbf{g'}$]{
%         \includegraphics[width = 0.46\linewidth]{figures/target_localization/localization_1.png}\label{subfig:localization_1}
%     }    
%     \subfigure[Alg. \ref{alg:localization-imaging} line 17: The refined estimate $\mathbf{\hat{r}}$ in line 17]{
%         \includegraphics[width = 0.46\linewidth]{figures/target_localization/localization_2.png}\label{subfig:localization_2}
%     }    
%     \caption{Illustrations of Algorithm \ref{alg:localization-imaging}, highlighting the locations of the key entities on the body's surface.}\label{fig:target_localization}
% \end{figure}
The matching score between an ultrasound image and a CT slice is 
% determined by first resampling them to have the same spacing, then convolving the HV segmentation mask of the ultrasound image with the binary HV segmentation label in the CT slice, and finally taking the maximum over the output matrix. The score 
is formally defined as follows: 
Let $Y$ and $\tilde{Y}$ be binary two-dimensional arrays with the same size. Let $Y\cap \tilde{Y}$ be the matrix resulting from entry-wise and-operation between $Y$ and $\tilde{Y}$. Let a translation map $t$ acting on $Y$ as an operation defined as follows: the map $t$ is parameterized by two \textbf{integer numbers} $\Delta x, \Delta y$. $X=t(Y)$ is a matrix that has the same dimensions as $Y$, where each entry $X_{ij}$ of $X$ is mapped either to $0$ or to an entry $Y_{kq}$ in $Y$, the exact formula being

$$
    \begin{bmatrix}k\\q\end{bmatrix} = 
    \begin{bmatrix}
            1 & 0 & \Delta x\\
            0 & 1 & \Delta y 
    \end{bmatrix} \begin{bmatrix}i\\j\\1\end{bmatrix}
$$
if the right hand side of the above is within the dimensions of $Y$, and $0$ otherwise. Applying $t$ to $Y$ results in shifting all the non-zero entries in $Y$ by offsets $(\Delta x, \Delta y)$.

Let $T(Y)=\{X: X= t(Y) \text{ for some translation map }t\}$ be the set of all possible translated versions of $Y$. The matching score in Algorithm \ref{alg:localization-imaging} is the optimal moving \textbf{Optimal Moving Intersection Area}(OMIA) between $Y,\tilde{Y}$, defined as
$$\text{OMIA}(Y,\tilde{Y})= \max_{X \in T(Y)}|\tilde{Y}\cap X|.$$ 

Note that OMIA can be easily computed by convolving $Y$ over $\tilde{Y}$, then taking the maximum over the output matrix. In the slice matching procedure disccused in Algorithm \ref{alg:localization-imaging}, $Y$ (HV segmentation from US) and $\tilde{Y}$ (HV segmentation from CT slice) do not have the same size. For OMIA to be properly defined, we pad $Y$ with zeros so that it ends up with the same size as $\tilde{Y}$, then compute OMIA between them to serve as the slice matching score in line 9 of Algorithm \ref{alg:localization-imaging}.

 

% %\bf{If you include a photo:}\vspace{-33pt}
% \begin{IEEEbiography}%[{\includegraphics[width=1in,height=1.25in,clip,keepaspectratio]{fig1}}]
% {Name}
% \end{IEEEbiography}


% %\bf{If you will not include a photo:}\vspace{-33pt}
% \begin{IEEEbiographynophoto}{Name2}

\vfill

\end{document}